%% file: main.tex
\documentclass[10pt,twocolumn,letterpaper]{article}

\usepackage{cvpr}              
\input{preamble}
\definecolor{cvprblue}{rgb}{0.21,0.49,0.74}
\usepackage[pagebackref,breaklinks,colorlinks,allcolors=cvprblue]{hyperref}

\usepackage{tcolorbox}
\usepackage{xcolor}
\usepackage{xspace}
\newcommand{\perception}{\textcolor[HTML]{2E75B6}{perception}\xspace}
\newcommand{\Perception}{\textcolor[HTML]{2E75B6}{Perception}\xspace}
\newcommand{\confidence}{\textcolor[HTML]{FFC000}{confidence}\xspace}
\newcommand{\Confidence}{\textcolor[HTML]{FFC000}{Confidence}\xspace}
\newcommand{\accuracy}{\textcolor[HTML]{548235}{accuracy}\xspace}
\newcommand{\Accuracy}{\textcolor[HTML]{548235}{Accuracy}\xspace}

\def\paperID{} 
\def\confName{CVPR}
\def\confYear{2026}

\title{Linking \Perception, \Confidence and \Accuracy in MLLMs}

\author{Yuetian~Du$^{1}$\thanks{Equal contribution.},~Yucheng~Wang$^{1}$\footnotemark[1],~Rongyu~Zhang$^1$, Zhijie~Xu$^4$, Boyu~Yang$^2$,\\Ming~Kong$^1$,~Jie~Liu$^{3}$\thanks{Corresponding Author.},~Qiang~Zhu$^{1}$\footnotemark[2] \\[2mm]
$^1$Zhejiang University~~~~$^2$Alibaba Group\\$^3$City University of Hong Kong~~~~$^4$University of Michigan \\ [1mm]
{\small Project:~\href{https://github.com/anotherbricki/CA-TTS}{https://github.com/anotherbricki/CA-TTS}}
}

\begin{document}
\maketitle
\input{sec/0_abstract}    
\input{sec/1_intro}

\input{sec/2_related_work}
\input{sec/3_method}

{
    \small
    \bibliographystyle{ieeenat_fullname}
    \bibliography{main}
}

\input{sec/X_suppl}

\end{document}

%% file: sec/0_abstract.tex
\begin{abstract}

Recent advances in Multi-modal Large Language Models (MLLMs) have predominantly focused on enhancing visual \perception to improve \accuracy. However, a critical question remains unexplored: \textbf{Do models know when they do not know?} Through a probing experiment, we reveal a severe \confidence miscalibration problem in MLLMs. To address this, we propose \textbf{Confidence-Driven Reinforcement Learning (CDRL)}, which uses original-noise image pairs and a novel confidence-based reward to enhance perceptual sensitivity and robustly calibrate the model's confidence. Beyond training benefits, calibrated confidence enables more effective test-time scaling as a free lunch. We further propose \textbf{Confidence-Aware Test-Time Scaling (CA-TTS)}, which dynamically coordinates Self-Consistency, Self-Reflection, and Visual Self-Check modules guided by confidence signals. An Expert Model acts in multiple roles (e.g., Planner, Critic, Voter) to schedule these modules and provide external verification. Our integrated framework establishes new state-of-the-art results with consistent 8.8\% gains across four benchmarks. More ablation studies demonstrate the effectiveness of each module and scaling superiority.


\end{abstract}

%% file: sec/1_intro.tex
\section{Introduction}

\begin{tcolorbox}[colback=gray!5, colframe=gray!40, boxrule=0.5pt]
\textit{``Ignorance more frequently begets confidence than does knowledge.''}

\vspace{0.8em}
\small \hfill --- Charles Darwin, \textit{The Descent of Man} (1871)
\end{tcolorbox}

\begin{figure}
    \centering
    \includegraphics[width=\linewidth]{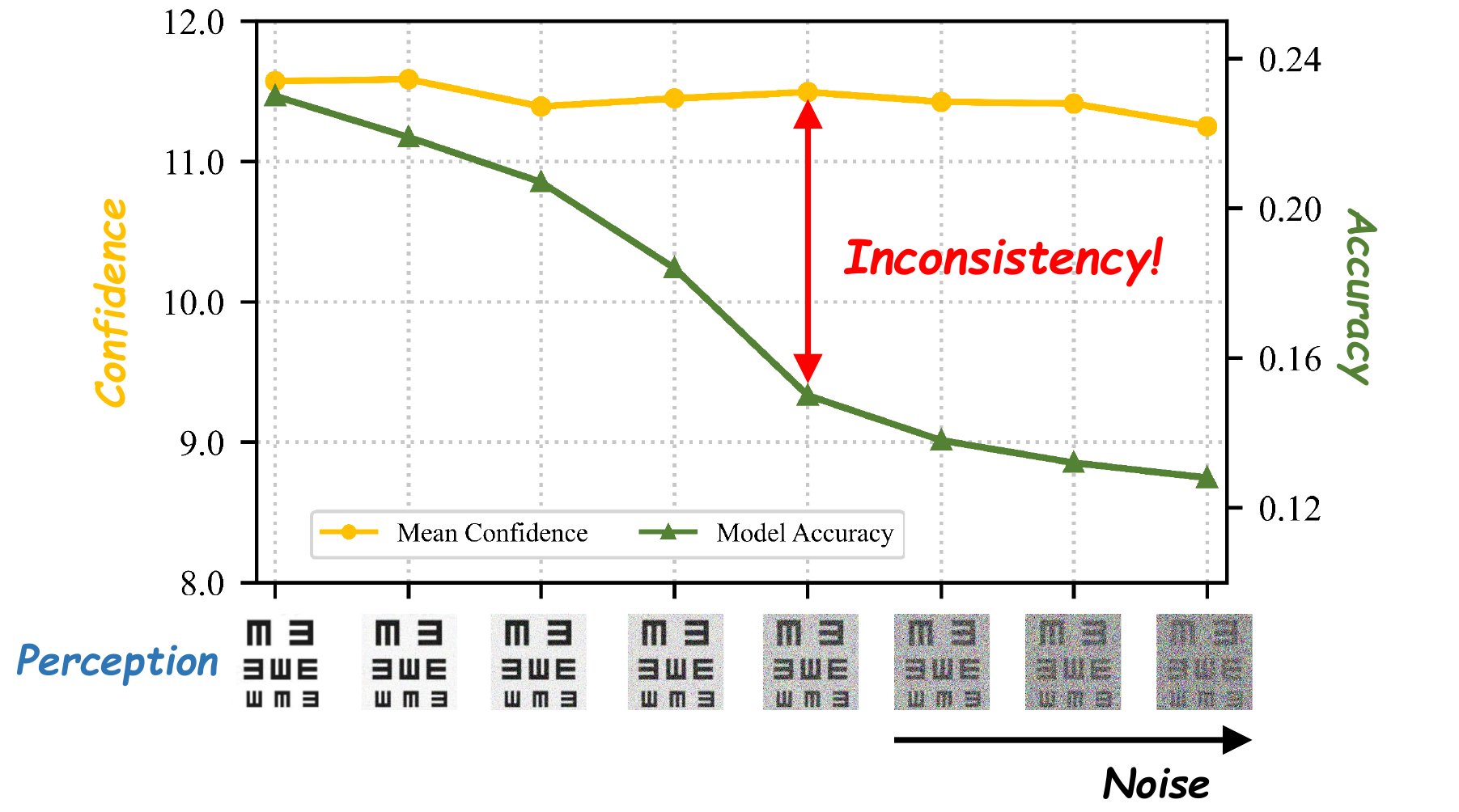}
    \caption{\textbf{Disconnection Between Model \Confidence and \Accuracy Under \Perception Degradation.} The X-axis ('Perception') shows the input image with progressively increasing noise. The plot demonstrates that while the \textit{\textbf{Mean Confidence}} remain highly stable (insensitive), the \textit{\textbf{Model Accuracy}} descends sharply, revealing a \textit{\textbf{significant gap} between the model's self-reported certainty and its actual performance as the visual input degrades.}}
    \label{fig:fig1}
\end{figure}

Recent advances \cite{luo2025sink,yu2024eliciting,huang2025revisiting,cai2025depthlm,han2025learning} in Multimodal Large Language Models (MLLMs) have focused on investigations into visual perception, ranging from optimizing visual data distributions~\cite{yu2024eliciting} and visual position encodings~\cite{huang2025revisiting} to refining visual instruction tuning~\cite{cai2025depthlm}. These efforts share a common goal: better visual \perception enable higher \accuracy. However, Darwin's paradox reminds us of an equally critical yet under-explored dimension: \textit{Does the model know when they do not know?}

To answer this question, we conduct a key probing experiment (Figure \ref{fig:fig1}). Specifically, we progressively add noise to key visual evidence containing critical information, and measure the model confidence and accuracy. If model truly relies on visual \perception, its \confidence and \accuracy should drop substantially when visual evidence disappears. However, we observe the opposite: \confidence remains surprisingly stable despite severe \perception degradation. This discrepancy exposes that MLLMs suffer from severe \textbf{confidence miscalibration}, which maintaining high \confidence even under \perception degradation.

This confidence miscalibration problem has been well studied in LLMs through uncertainty~\cite{xiong2023uncertainty,ji2025calibratingverbaluncertaintylinear} and logits~\cite{kang2025scalablebestofnselectionlarge,fu2025deep} estimation for each textual output token and obtain satisfactory performance. However, these works fundamentally misalign with MLLMs' visual perception due to granularity mismatch. While LLMs calibrate confidence at \textit{individual token granularity}, visual perception in MLLMs manifests holistically \textit{across the entire response}. To solve this problem, we calculate the confidence for entire response as the mean negative log-probability across all output tokens. Building on this, we propose a \textbf{Confidence-Driven Reinforcement Learning (CDRL)} approach that explicitly rewards perception-confidence alignment. This enables MLLMs to develop calibrated confidence that accurately reflects their visual understanding, bridging the gap between what they see and what they claim to know.


A free lunch for this confidence calibration is its direct applicability to test-time scaling with considerable improvement. The calibrated confidence naturally serves as a reliability indicator, enabling models to identify uncertain predictions that warrant additional reasoning effort. To fully leverage this advantage, we further propose \textbf{Confidence-Aware Test-Time Scaling (CA-TTS)}, which coordinates three proposed synergistic modules guided by confidence signals. \textit{Self-Consistency} employs confidence-weighted voting combined with expert model calibration to aggregate multiple reasoning paths. \textit{Self-Reflection} leverages expert-generated critiques to refine low-confidence predictions. \textit{Self-Check} validates visual grounding through contrastive decoding between original and noised images. By dynamically routing to appropriate modules based on confidence levels, CA-TTS achieves substantial accuracy gains while maintaining computational efficiency.

Our contributions are summarized as follows:
\begin{itemize}
    \item We present \textbf{the first systematic investigation} of visual perception-aware confidence calibration in MLLMs.
    
    \item We propose \textbf{Confidence-Driven RL from Vision (CDRL)}, a calibration training method with specialized confidence rewards to enhance perceptual sensitivity.
    
    \item We show CDRL's calibrated confidence enables test-time scaling as a free lunch, enabling our \textbf{Confidence-Aware Test-Time Scaling (CA-TTS).}
    
    \item Our integrated framework achieves new state-of-the-art results, significantly outperforming baselines with consistent \textbf{8.8\%} overall gain on four benchmarks.
\end{itemize}

%% file: sec/2_related_work.tex
\section{Related Work}
\subsection{Visual Perception Studies of MLLMs}

The perceptual capabilities of MLLMs initially benefited from the integration of independently pre-trained models. CLIP \cite{radford2021learningtransferablevisualmodels} achieved alignment between visual and textual representations via contrastive learning. Subsequent models, such as LLaVA \cite{Liu2023VisualIT} and Qwen-VL \cite{Bai2023QwenVL}, leveraged more powerful architectures and higher-quality data to enhance visual understanding and instruction-following abilities.

\begin{figure*}[t]
    \centering
    \includegraphics[width=0.95\linewidth]{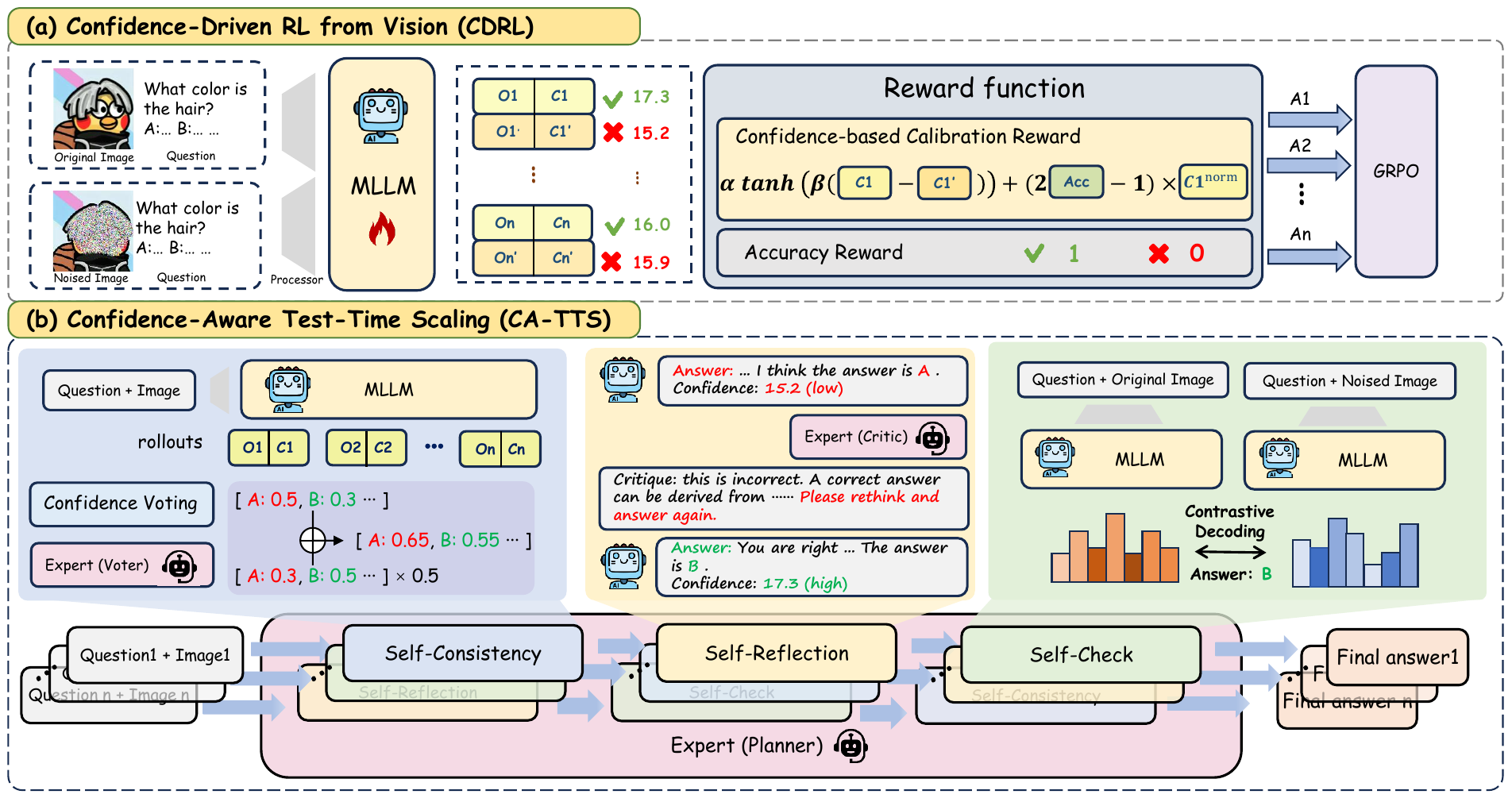}
    \caption{\textbf{Framework Overview.} The upper panel \textbf{(a) }illustrates how original-noise image pairs are used to optimize the model via Reinforcement Learning, driven by a \textbf{Confidence-based Calibration Reward} and a Accuracy Reward. The bottom panel \textbf{(b)} shows the adaptive \textbf{Confidence-Aware Test-Time Scaling (CA-TTS)} system, where an Expert Model acts as a \textbf{Planner, Voter, and Critic} to coordinate the \textbf{Self-Consistency, Self-Reflection, and Self-Check} modules, which collaborate to produce the final answer.}
    \label{fig:fig2}
\end{figure*}

Key factors influencing perceptual capabilities have been identified. In the realm of visual processing, MLLMs demonstrate a tendency to over-focus on a few visual tokens \cite{Shen2025Sink}. Regarding in-context learning, they still show difficulty in effectively leveraging visual cues for fine-grained reasoning \cite{Li2025Context}. Positional encoding methods also present persistent challenges, such as modal confusion and inadequate multi-scale representation \cite{Wang2025Revisiting}. Furthermore, reinforcement learning based on preference optimization has been shown to significantly enhance performance on vision-intensive tasks compared to SFT, while also strengthening the vision encoder's representational capabilities \cite{song2025rlmakesmllmsbetter}.

Inspired by Reinforcement Learning from Human Feedback (RLHF), research has begun to explore its application in MLLM visual perception. Incorporating human preference data has been shown to enhance performance \cite{Wang2024b} and mitigate hallucinations \cite{Yang2025c, Yu2024, Fu2025b}. For instance, \cite{Liu2024LLaVA1_6} utilized AI feedback to construct reward models, improving output faithfulness. Methods such as DPO have also been employed to capture subtle visual differences \cite{Xie2024}. While these approaches have demonstrably improved perceptual abilities, the role of \textbf{confidence} has been seldom investigated.

\subsection{Calibration for MLLMs}
Multi-modal Large Language Models (MLLMs) suffer from severe \textbf{systemic miscalibration} when evaluating their own outputs, with overconfidence leading to a significant gap between reported confidence and actual accuracy. Unlike the relatively mature research on LLM calibration \cite{geng2023survey,liu2023trustworthy,xiong2023uncertainty,ni2024retrieval,tian2023calibration,tao2024reinforce,wang2024verb,kang2025scalablebestofnselectionlarge}, work on MLLM calibration is nascent, facing the core challenge of addressing the unique impact of the visual component.

This issue initially gained traction in high-risk domains like clinical diagnostics and autonomous driving \cite{DuYue_Confidence_MICCAI2025, kriz2025prompt4trustreinforcementlearningprompt}, where solutions focused on Multi-round Interrogation or RL-Prompting. Furthermore, other research \cite{shen2025exposingmitigatingcalibrationbiases} has explored training-free methods, applying them to in-context learning (ICL) for medical image classification. More recently, as research identifies hallucinations as an extreme form of calibration failure, advancements such as visual contrastive decoding \cite{leng2023mitigatingobjecthallucinationslarge,park2024conviscontrastivedecodinghallucination} and DPO-based alignment \cite{compagnoni2025mitigatinghallucinationsmultimodalllms} have emerged. However, these efforts largely remain incremental extensions of LLM calibration methods. They fail to adequately address how core concepts like visual perception fundamentally impacts calibration outcomes---the key differentiator from LLM calibration.

\subsection{Test-Time Scaling Strategies}

As enthusiasm for scaling computation during pre-training wanes, Test-Time Scaling (TTS), also known as Test-Time Compute, has emerged as a significant research focus \cite{zhang2025surveytesttimescalinglarge,alomrani2025reasoningbudgetsurveyadaptive,ji2025surveytesttimecomputeintuitive}. TTS aims to establish a paradigm where increasing computational expenditure during inference yields consistent performance improvements. This approach has gained traction due to its potential for greater generalization and flexibility compared to the standard pre-training and fine-tuning framework. Early practices, such as CoT-prompting \cite{wei2023chainofthoughtpromptingelicitsreasoning}, initiated this line of inquiry. Subsequently, TTS research diverged into two primary branches: (1) \textbf{Parallel Scaling}, represented by methods like Self-Consistency \cite{li2025stesttimescaling}, which combines multi-sampling and majority voting; and (2) \textbf{Sequential Scaling}, exemplified by approaches such as Self-Refine \cite{madaan2023selfrefineiterativerefinementselffeedback} and STaR \cite{zelikman2022starbootstrappingreasoningreasoning}. The integration of these two branches has led to the development of tree-structured \textbf{Hybrid Scaling} strategies like ToT \cite{yao2023treethoughtsdeliberateproblem} and MCTS-based methods \cite{lin2025cmctsconstrainedmontecarlo}, which enhance policy robustness. Furthermore, large-scale RL-based inference models, spearheaded by Openai-o1 \cite{openai2024openaio1card} and Deepseek-R1 \cite{deepseekai2025deepseekr1incentivizingreasoningcapability}, can be viewed as a novel \textbf{Internal Scaling} paradigm.

More recently, research has built upon these foundational paradigms to conduct cutting-edge explorations. For instance, TTRL \cite{zuo2025ttrltesttimereinforcementlearning} ingeniously adapts these concepts to Test-Time Training (TTT), enhancing a model's ability to learn from unlabeled data. s1 \cite{muennighoff2025s1simpletesttimescaling} demonstrated that TTS techniques could achieve performance comparable to large models trained on massive datasets using only about 1k samples. Highlighting its potential, other work has shown a 3B model outperforming a 405B model through TTS \cite{liu20251bllmsurpass405b}. Deepconf \cite{fu2025deep} also achieved substantial improvements in mathematical reasoning solely by using confidence scores for TTS. Concurrently, efforts have begun to investigate the potential of TTS in the multimodal domain. For example, researchers have proposed Test-Time Reranking (TTR) \cite{hsu2025testtimescalingstrategiesgenerative}, which uses expert model confidence to refine the original model's probability distribution. However, robust Multimodal TTS that fundamentally addresses the role of visual components and framework robustness remains a significant research gap.

%% file: sec/3_method.tex
\section{Method}
\subsection{Framework Overview}
To address the Perceptual Bluntness Problem in visual reasoning, we propose an innovative framework (Figure \ref{fig:fig2}) built upon two core components. \textbf{Confidence-Driven RL from Vision (CDRL)} enhances perceptual sensitivity and calibrates confidence using GRPO and original-noise image pairs. Based on the calibrated confidence, \textbf{Confidence-Aware Test-Time Scaling (CA-TTS)} employs an adaptive strategy where an Expert Planner coordinates multiple decoupled reasoning modules to ensure a robust, final answer.

\subsection{Confidence-Driven RL from Vision (CDRL)}
\label{sec:3.2}

\subsubsection{Preliminaries}

\textbf{High-Quality Data Filtering.} To construct the high-quality training set $D_{\text{RL}}$, we first aggregate $D_{\text{source}}$ from six public benchmarks: three for mathematical reasoning\cite{zhang2024mathversedoesmultimodalllm,zou2025dynamathdynamicvisualbenchmark,xiao2024logicvistamultimodalllmlogical} and three general-purpose VQA datasets \cite{xai_realworldqa,liu2024mmbenchmultimodalmodelallaround,yue2025mmmuprorobustmultidisciplinemultimodal}. We use an LLM-based pipeline to filter this pool for quality, difficulty, and diversity, yielding $D_{\text{Filtered}}$ with 1936 data. Detailed information about filtering procedure and dataset can be found in Appendix \ref{app:b}.

\label{sec:3.2}
\noindent \textbf{Noised Image Generation.} To sensitize the model to perturbations, we augment $D_{\text{Filtered}}$ by using CLIP attention maps to apply a noise function $\mathcal{G}_{\text{noise}}$, generating a perturbed image $i'$ for each original $i$. This creates the final training set $D_{\text{RL}}$, containing paired tuples of $((i, i'), q, a)$. This dataset is fundamental for the following resource-friendly GRPO algorithm, which aims to enhance the model's perceptual sensitivity and robust self-calibration capabilities.

\noindent \textbf{Group Relative Policy Optimization.} We employ a policy model $\pi_{\theta}$ and an initial reference model $\pi_{\text{ref}}$. In each training loop, we sample an image pair $(i, i')$ and question $q$ from $D$. The policy then generates $k$ candidate output pairs, $(o_1, o_1'), \dots, (o_k, o_k')$, where $o_j$ is the reasoning trajectory for the original image $i$. The GRPO algorithm optimizes $\pi_{\theta}$ by maximizing the objective:
\begin{equation}
\begin{split}
J_{GRPO}(\theta) = \mathbb{E}_{(i, i', q) \sim D} \left[ \frac{1}{k} \sum_{j=1}^{k} R(i, i', q, o_j) \right] \\
- \beta D_{KL}(\pi_{\theta} \| \pi_{\text{ref}}).
\end{split}
\label{eq:grpo_obj}
\end{equation}

In practice, optimization uses advantage estimation. Each of the $k$ outputs $o_j$ is scored with a total reward $r_j$. Its advantage $A_j$—the reward normalized against the group's average—reflects its relative quality. High-advantage paths are then up-sampled, while low-advantage paths are suppressed.

Our total reward score $r_j$, used for the advantage estimation, comprises three components: the output accuracy reward $R_{\text{Output}, j} = \mathbb{I}(\text{GT} \subseteq a_j)$ (where $a_j$ is the final answer, $\text{GT}$ is the ground truth, and $\mathbb{I}$ is the indicator function), a formatting reward $R_{\text{Format}, j}$ to ensure structural correctness, and our novel \textbf{Confidence-based Calibration Reward} $R_{\text{Conf}, j}$. The final reward is:
\begin{equation}
    r_j = R_{\text{Conf}, j} + R_{\text{Output}, j} + R_{\text{Format}, j}
\end{equation}

\subsubsection{Confidence-based Calibration Reward}

To address both perceptual insensitivity and poor confidence calibration, we design the \textbf{Confidence-based Calibration Reward}, $R_{\text{Conf}, j}$. This reward aims to enhance perceptual sensitivity using the image pair $(i, i')$ and their outputs $(o_j, o_j')$, while also promoting good confidence calibration.

\label{sec:logits}
To compute this reward, we first need to define the model's output confidence $C$. At each generation step $t$, we compute the Negative Mean Log-Probability (NMLP) of the model's output logits $L$. First, the logits $L$ are converted to log-probabilities: $\mathbf{lp} = \text{LogSoftmax}(L)$. We then select the top $k$ highest log-probability values (denoted $\log p_{(i)}$) to define the token's confidence $\text{Conf}_{\text{token}}$. For a complete sequence $o$ composed of $T$ tokens, its total confidence $C$ is the arithmetic mean of all token confidences:
\begin{equation}
C = \frac{1}{T} \sum_{t=1}^{T} \text{Conf}_{\text{token}_t}, \  \text{where}\  \text{Conf}_{\text{token}} = - \frac{1}{k} \sum_{i=1}^{k} \log p_{(i)}.
\label{eq:conf_combined}
\end{equation}
A \textbf{lower} NMLP value indicates a \textbf{sharper} probability distribution, signifying higher certainty.


Our $R_{\text{Conf}}$ reward combines these objectives. The perception goal uses the raw confidence difference $\Delta C = C_j - C_j'$ (from outputs $o_j, o_j'$) to encourage sensitivity. The calibration goal links the binary accuracy $R_{\text{Output}, j}$ (which is 0 or 1, defined previously) with the normalized confidence $C_{j}^{norm}$ from the original image output. This combined mechanism encourages the model to be confident when correct and unconfident when wrong, while simultaneously being sensitive to input perturbations.

The final \textbf{Confidence-based Calibration Reward} $R_{\text{Conf}, j}$ is formulated as:
\begin{equation}
R_{\text{Conf}, j} = \underbrace{\alpha \tanh(\beta * \Delta C)}_{\text{Perception Term}} + \underbrace{(2 \cdot R_{\text{Output}, j} - 1) \cdot C_{j}^{norm}}_{\text{Calibration Term}},
\label{eq:r_Conf_combined} 
\end{equation}
where the first term rewards a large confidence change $\Delta C$ and the second term (which simplifies to $+C_j$ if correct and $-C_j$ if incorrect) rewards proper calibration.

\textit{Note:} To avoid contaminating the training trajectory, the calibration term and its associated gradient updates are only computed based on the output $o_j$ (from the original image) and its confidence $C_j$, not the perturbed output $o_j'$.

\subsection{Confidence-Aware Test-Time Scaling (CA-TTS)}

In this section, we introduce our adaptive multi-module TTS framework. This framework comprises three core modules driven by confidence and enhanced by integrating visual and textual elements: \textbf{Self-Consistency, Self-Reflection,} and \textbf{Self-Check}. Furthermore, an Expert Model is incorporated to schedule the interactions between these modules or to participate deeply within specific tasks, operating under different designated roles.

\subsubsection{Self-Consistency}

Instead of merely relying on multi-sampling and majority voting, our self-consistency is based on a confidence-driven, Expert Model-guided approach.

First, for each input sample (Image $i$, Question $q$), we collect $n$ samples, gathering their corresponding Chain-of-Thought ($CoT$), Answer ($A$), and Confidence ($C$) sequences:
\begin{equation}
S = \{ (CoT_i, A_i, C_i) \}_{i=1}^n.
\end{equation}

Subsequently, leveraging the reliable confidence and visual perception capabilities obtained in the first stage, and inspired by prior work \cite{fu2025deep}, we employ confidence-weighted majority voting. We experimented with various confidence quantification methods (e.g., tail confidence, min confidence) and found mean confidence to be the most effective.

Specifically, we aggregate the confidence scores for each candidate answer $k$ across all samples. This serves as the aggregation and calibration of the model's internal confidence. The result of this internal voting, $V_{internal}$, is a dictionary mapping candidate options $k$ to their weighted vote counts:
\begin{equation}
V_{internal}[k] = \sum_{i=1}^n C_i \cdot \mathbb{I}(A_i = k),
\end{equation}
where $\mathbb{I}$ is the indicator function (1 if $A_i = k$, 0 otherwise). Next, we combine this internal calibration with an external calibration step. We first extract the set of all unique candidate answers from our initial samples to form a candidate list $L_{candidates}$. We then provide the image $i$, question $q$, and this candidate list $L_{candidates}$ to the Expert Model (now in the \textbf{Voter} role, the prompt $\text{P}_{\text{voter}}$ is shown in Appendix \ref{app:a}). The expert $M_{expert}^{Voter}$ is designated to \textbf{verbally }output its internal confidence for each option, producing a normalized confidence list $C_{expert} = [c_1, \dots, c_{|L|}]$ such that $\sum_j c_j = 1$. We then apply a voting weight $\tau_1$. This weight is multiplied by the expert's normalized confidence $c_k$ for a given answer $k$. This external vote is added to the normalized internal voting score $V_{internal}^{norm}[k]$ to yield the final vote dictionary $V_{final}$:
\begin{equation}
V_{final}[k] = V_{internal}^{norm}[k] + \tau_1 \cdot c_k, \quad \forall k \in L_{candidates}.
\end{equation}

In this manner, the \textbf{Voter Expert} participates in the calibration process, providing a secondary adjustment opportunity to the base model's internal calibration, thus making the entire self-consistency process more robust.

\subsubsection{Self-Reflection}

We employ the Expert Model as a \textbf{Critic} ($M_{expert}^{Critic}$) to generate critiques that guide the base model in reconsidering its initial reasoning. In this phase, we provide the original image $i$ and question $q$ to the expert. Using a specific prompt ($\text{P}_{\text{critique}}$, see Appendix \ref{app:a}), the expert is directed to generate a corresponding critique ($Crit$) on the problem.
\begin{equation}
\begin{split}
Crit = M_{expert}^{Critic}(i, q, \text{P}_{critique}).
\end{split}
\end{equation}

After obtaining the $Crit$, we use this critique to prompt the base model ($M_{base}$) to reconsider its reasoning and generate a new reflected Chain-of-Thought $CoT_{reflect}$ and its corresponding answer $A_{reflect}$:
\begin{equation}
(CoT_{reflect}, A_{reflect}) = M_{base}(i, q, Crit).
\end{equation}

This reflected answer $A_{reflect}$ is then added to the final vote tally $V_{final}$ by incrementing its score by a weight of $\tau_2$. If $A_{reflect}$ is a new answer not previously in $V_{final}$, it is first initialized with this score.

\subsubsection{Self-Check}

This module shifts the focus from text-based checks to self-examination at the visual level. Using the same image pair construction method as in Section \ref{sec:3.2}, we create an original-noise image pair ($i, i'$) for the test image. As noted in Section \ref{sec:logits}, logits can be considered a precursor form of confidence. Therefore, inspired by \cite{leng2023mitigatingobjecthallucinationslarge}, we utilize a more fundamental, confidence-driven method by applying Visual Contrastive Decoding (VCD) to adjust the output. This process decodes the answer by contrasting the log probabilities of generating answer $y$ from the original image $i$ and the noisy image $i'$:
\begin{equation}
\begin{split}
\log P_{VCD}(y | i, q) &= (1+\alpha)\cdot\log P_{\theta}(y | i, q) \\
&\quad - \alpha \cdot \log P_{\theta}(y | i', q),
\end{split}
\end{equation}
where $\alpha$ is a hyperparameter controlling the strength of the contrast.

This module does not directly involve the Expert Model. The answer obtained from VCD decoding, $A_{check}$, is added to the $V_{final}$ dictionary by incrementing its score by a voting weight $\tau_3$.

\begin{table*}[htbp]
\centering
\caption{\textbf{Evaluation Results on Visual Reasoning Benchmarks.} Abbreviations: OE (Open-Ended), MC (Multi-Choice), PE (Perception), RE (Reasoning), STEM (Science, Technology, Engineering, and Mathematics), HASS (Humanities, Arts, and Social Sciences), and ALL (Overall). Best results are \textbf{bold}. }
\resizebox{0.9\linewidth}{!}{
\begin{tabular}{l ccc ccc ccc ccc}
\hline
& \multicolumn{6}{c}{\textbf{Math Reasoning}} & \multicolumn{6}{c}{\textbf{General VQA}} \\
\cmidrule(lr){2-7} \cmidrule(lr){8-13}
\textbf{Model/Dataset} & \multicolumn{3}{c}{\textbf{Math-Vista}$_{testmini}$} & \multicolumn{3}{c}{\textbf{Math-Vision}$_{test}$} & \multicolumn{3}{c}{\textbf{MMStar}$_{test}$} & \multicolumn{3}{c}{\textbf{MMMU}$_{val}$} \\
\cmidrule(lr){2-4} \cmidrule(lr){5-7} \cmidrule(lr){8-10} \cmidrule(lr){11-13}
& OE & MC & ALL & OE & MC & ALL & PE & RE & ALL & STEM & HASS & ALL \\
\hline
\multicolumn{13}{l}{\textit{Training Free Baselines (Qwen2.5-VL-7B)}} \\
\cline{1-13}Pass@1 & 62.3 & 66.8 & 64.7 & 24.2 & 21.7 & 23.0 & 58.4 & 68.2 & 60.2 & 40.1 & 60.7 & 48.8 \\
Majority Voting & 68.4 & 73.7 & 69.8 & 26.2 & 33.2 & 30.1 & 63.6 & 77.0 & 69.0 & 53.2 & 62.5 & 57.5 \\
Deepconf & 69.3 & 74.7 & 70.7 & 26.4 & 32.3 & 29.6 & 58.2 & 71.0 & 61.1 & 51.5 & 62.5 & 56.1 \\
\hline
\multicolumn{13}{l}{\textit{Training Based Framework}} \\
\cline{1-13}
DreamPRM(InternVL-2.5-8B) & -& -& 68.9 & -& -& 22.1 &- & -& 62.3 &- &- & 61.4 \\
R1-Onevision(Qwen2.5-VL-7B) & 61.7 & 66.2 & 64.1 & 20.8 & 37.8 & 29.9 & 60.8 & 66.9 & 63.7 & 51.6 & 59.3 & 55.3 \\
VL-Rethinker(Qwen2.5-VL-7B) & 65.3 & 81.7 & 74.1 & 22.1 & 39.0 & 30.7 & 60.7&66.5 &63.4 & 51.6 & 60.7 & 55.6 \\
We-Think(Qwen2.5-VL-7B) & 65.7 & 80.1 & 73.3 & 20.9 & 37.4 & 29.7 & 62.6 & 70.1 & 65.1 & 50.8 & 62.5 & 55.7 \\
\textbf{Ours (Qwen2.5-VL-7B)} & \textbf{74.2} & \textbf{84.7} & \textbf{79.5} & \textbf{38.4} & \textbf{45.6} & \textbf{42.4} & \textbf{67.8} & \textbf{77.2} & \textbf{71.3} & \textbf{59.9} & \textbf{73.5} & \textbf{66.3} \\
\hline
\end{tabular}
}
\label{tab:model_performance_full}
\end{table*}

\subsubsection{Expert Planning}

In addition to its {Voter} and {Critic} roles, the Expert Model also functions as a {Planner}, $M_{expert}^{Planner}$, responsible for module scheduling. Before inference, the planner analyzes the input $(i, q)$ and outputs a scheduling order $\pi$. This order is a permutation of the three modules (Self-Consistency $M_{sc}$, Self-Reflection $M_{sr}$, and Self-Check $M_{sk}$), ensuring that each module is used exactly once.

This adaptive scheduling is feasible because the three modules are fully decoupled and order-insensitive. The final output of every module is simply a contribution to the shared voting dictionary $V_{final}$. If a module is executed first, it initializes the contributions to an empty $V_{final}$. This design ensures the flexibility and robustness of our system.





\section{Experiments}

\subsection{Experimental Setup}






\subsubsection{Model Baselines}

Our framework uses Qwen2.5-VL-7B-Instruct \cite{qwen2.5-VL} as the base model in a two-stage process. First, the \textbf{CDRL} training stage enhances the model's perceptual sensitivity. Second, the \textbf{CA-TTS} inference stage uses this trained model as the reasoning agent. In the \textbf{CA-TTS} framework, we employ Gemini-2.5-Pro \cite{comanici2025gemini} as the Expert Model. Its role is to adaptively schedule the three core TTS modules and provide verification feedback.

\subsubsection{Evaluation Benchmarks}

We evaluated our framework's effectiveness and robustness on key benchmarks covering image-based mathematical reasoning and general multimodal reasoning. The benchmarks include:
\begin{itemize}
    \item \textbf{Math-Vista \cite{lu2023mathvista}:} A comprehensive mathematical vision reasoning benchmark, integrating 28 existing multimodal datasets and 3 newly created ones.
    \item \textbf{Math-Vision \cite{wang2024measuring}:} A high-quality multimodal math reasoning benchmark containing 3040 samples, spanning 16 different mathematical disciplines.
    \item \textbf{MMStar \cite{chen2024mmstar}:} A vision-indispensable general-purpose multimodal benchmark, containing 1500 meticulously human-curated samples.
    \item \textbf{MMMU \cite{yue2024mmmu}:} A massive multidisciplinary multimodal benchmark designed to evaluate model performance across 30 subject areas and 183 subfields.
\end{itemize}

\subsubsection{Baselines}

We compare our method (\textbf{CDRL + CA-TTS}) against two categories of baselines to validate its superiority:
\begin{enumerate}
    \item \textbf{Training-Free Baselines:} Includes various training-free Test-Time Scaling (TTS) strategies, such as Majority Voting \cite{li2025revisiting} and Deepconf \cite{fu2025deep}. To ensure a fair comparison, all Training-Free methods were reproduced on the same base model (Qwen2.5-VL-7B-Instruct).
    \item \textbf{Training Baselines:} Includes various models trained on visual reasoning tasks, such as DreamPRM \cite{cao2025dreamprm}, R1-Onevision \cite{yang2025r1}, VL-Rethinker \cite{wang2025vlrethinkerincentivizingselfreflectionvisionlanguage} and WeThink \cite{yang2025wethinkgeneralpurposevisionlanguagereasoning}.
\end{enumerate}

\begin{table}[t]
\centering
\caption{\textbf{Ablation Study for CDRL and CA-TTS.} Best results are \textbf{bold}, second-best are \underline{underlined}.}
\resizebox{0.8\linewidth}{!}{
\begin{tabular}{l p{1cm} p{1cm} p{1cm}}
\hline
& \multicolumn{3}{c}{\textbf{Math-Vision}$_{test}$} \\
\cmidrule(lr){2-4}
\textbf{Setting} & OE & MC & ALL \\
\hline
Training-Free & 24.24 & 21.71 & 22.96 \\
CDRL & 18.46 & 34.16 & 26.38 \\
CA-TTS & \underline{37.99} & \underline{42.95} & \underline{37.99} \\
{CDRL+CA-TTS} & \textbf{38.44} & \textbf{45.60} & \textbf{42.35} \\
\hline
\end{tabular}}
\label{tab:table2}
\end{table}




\subsection{Implementation Details}

\noindent \textbf{CDRL.} We performed full-parameter fine-tuning on 8$\times$H100 141GB GPUs using bfloat16 mixed precision and a batch size of 2. We generated 4 rollout pairs per sample, though rollouts from noised-image inputs did not participate in the gradient update.

\noindent \textbf{CA-TTS.} We generated 8 parallel inference samples per question, using Temperature $T=1.0$ and $top \mbox{-} k=40$ to encourage diversity. All module voting weights were set equally ($\tau_1 = \tau_2 = \tau_3 = 0.5$). For the Self-Check module, VCD hyperparameters were set to $\alpha = 0.5$ and $\beta = 0.1$. The Expert Voter was allowed a maximum of 3 retries to ensure reliable confidence output.













\subsection{Main Results}

\noindent\textbf{a) Superior Performance over Training-Based Baselines.} As shown in Table 1, our proposed method (Ours) achieves state-of-the-art performance across all four visual reasoning benchmarks. Specifically, our model (Ours (Qwen2.5-VL-7B)) reaches ALL scores of 79.5\%, 42.4\%, 71.3\%, and 66.3\% on Math-Vista, Math-Vision, MMStar, and MMMU, respectively. More importantly, compared to other advanced training-based methods like VL-Rethinker, our model demonstrates stronger performance on all benchmarks—for instance, achieving 0.7\% higher on Math-Vista and 0.7\% higher on MMMU. This fully demonstrates the effectiveness and superior generalization capability of our proposed framework.

\noindent\textbf{b) Validating the Free Lunch of Calibrated Confidence.} This significantly outperforms training-free baselines, including Pass@1 and Majority Voting. This result validates the free lunch concept introduced earlier: the calibrated confidence from our training phase is directly and effectively translated into considerable performance improvements at inference time via our \textbf{Confidence-Aware Test-Time Scaling (CA-TTS)} framework.

\begin{table}[tbp]
\centering
\caption{\textbf{Performance of Different Expert Models on Math-Vision.} OE (Open-Ended), MC (Multi-Choice). The results on other datasets can be found in Appendix \ref{app:c}.}
\resizebox{0.75\linewidth}{!}{
\begin{tabular}{l ccc}
\hline
& \multicolumn{3}{c}{\textbf{Math-Vision}$_{testmini}$} \\
\cmidrule(lr){2-4}
\textbf{Expert Model} & OE & MC & ALL \\
\hline
Majority Voting & 22.09 & 31.30 & 27.65 \\
Qwen-2.5-VL-7B & 30.91 & 34.03 & 32.57 \\
Qwen-2.5-VL-72B & 28.52 & 37.70 & 34.21 \\
Qwen-VL-Max  & 35.71 & 36.13 & 35.97 \\
GPT-5  & \underline{38.94} & \underline{43.93} & \underline{41.45} \\
Gemini-2.5-Pro & \textbf{46.62} & \textbf{42.41} & \textbf{43.75} \\
\hline
\end{tabular}}
\label{tab:table3}
\end{table}

\subsection{Ablation Studies}

We conducted a series of ablation studies and conclude four insights as shown in following:

\label{4:scaling}
\noindent \textbf{a) CDRL and CA-TTS contribute independently and synergistically.} In Table \ref{tab:table2}, we analyze the contributions of different components. Here, Training-Free corresponds to the Pass@1 baseline from Table 1. (1) Using the \textbf{CDRL} alone provides a moderate performance boost over the baseline (e.g., from 48.8\% to 52.2\% on MMMU). (2) Using the \textbf{CA-TTS} alone yields a significant leap in performance (e.g., jumping from 64.7\% to 77.8\% on Math-Vista). (3) By combining both (\textbf{CDRL+CA-TTS}), our full model achieves the best performance across all benchmarks (e.g., 42.4\% on Math-Vision). This suggests that the \textbf{CDRL} stage provides a better policy or model state for the \textbf{CA-TTS} stage, and the combination is most effective.

\label{sec:self}
\noindent \textbf{b) This framework can be generalized to different expert models.} To investigate the generalizability and scalability of our framework, we applied it to a series of different expert models. As shown in Table \ref{tab:table3}, we evaluated performance using Majority Voting as a comparison baseline and found better performance \textit{even Qwen-2.5-VL-7B itself serves as an expert model}. The evaluation also extended to other powerful foundational models, including Qwen-2.5-VL-72B, Qwen-VL-Max, GPT-5, and Gemini-2.5-Pro. The experiments were conducted on the Math-Vision$_{testmini}$ dataset. This diversity of experts aims to verify that our method is robust and not highly dependent on the performance of any single expert model. This further highlights that our framework, particularly through its self-calibration capabilities, can consistently and effectively enhance the reasoning capabilities of models with different scales and architectures.

\begin{figure}
    \centering
    \includegraphics[width=0.8\linewidth]{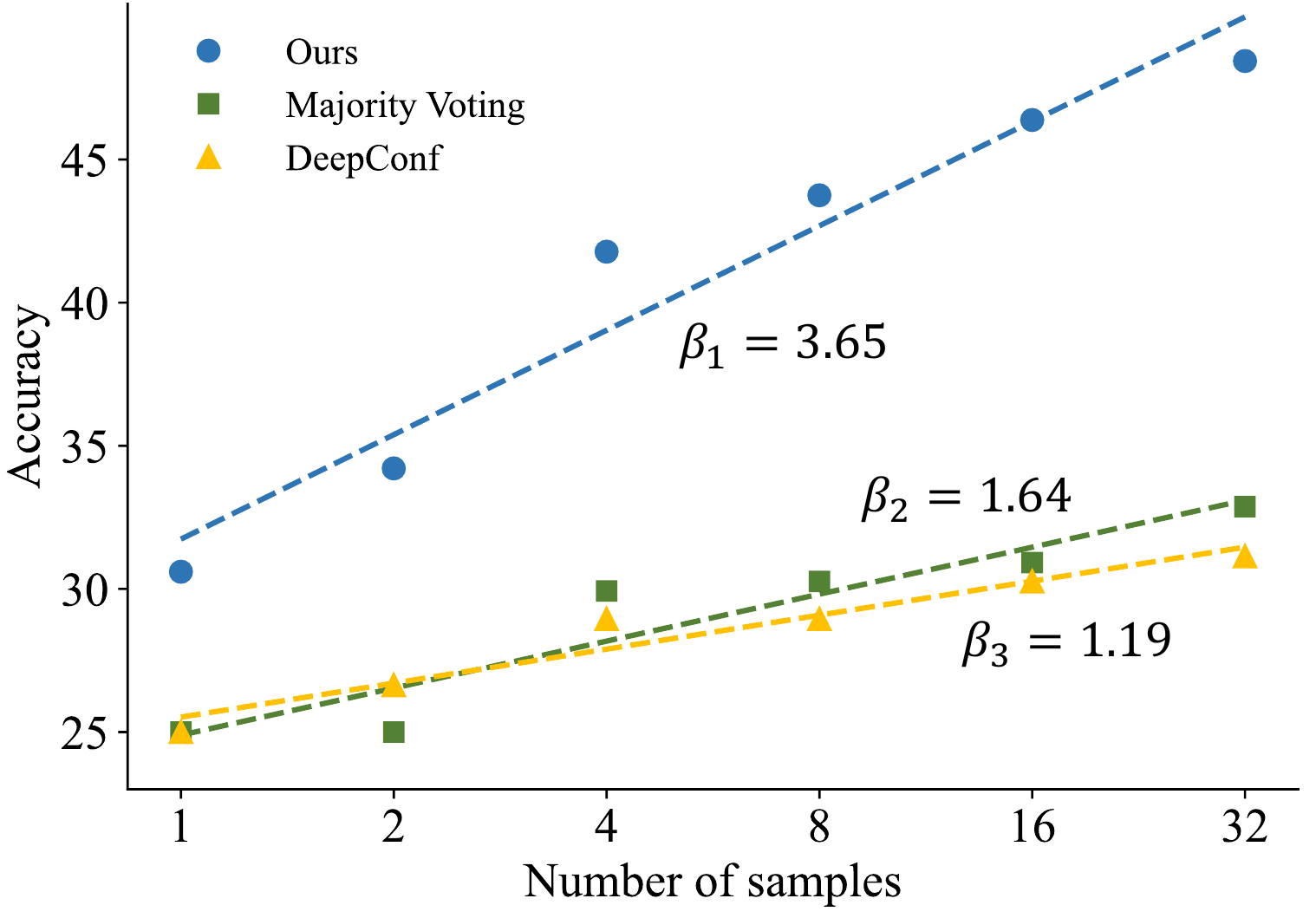}
    \caption{\textbf{Test-time scaling comparison on Math-Vision.} Accuracy vs. number of samples for our \textbf{CA-TTS} (blue), Majority Voting (green), and DeepConf (yellow). The slope of our method ($\beta_1 = 3.65$) is 2.2-3.1$\times$ steeper than baselines ($\beta_2 = 1.64$, $\beta_3 = 1.19$), demonstrating superior scaling potential with calibrated confidence.}
    \label{fig:fig3}
\end{figure}

\noindent \textbf{c) Test-time scaling is enhanced with calibrated confidence.} As shown in Figure~\ref{fig:fig3}, our \textbf{CA-TTS} method demonstrates superior scaling properties compared to Majority Voting and DeepConf baselines. The key advantage is the substantially steeper scaling slope: our method achieves $\beta_1 = 3.65$, which is 2.2$\times$ and 3.1$\times$ higher than Majority Voting ($\beta_2 = 1.64$) and DeepConf ($\beta_3 = 1.19$), respectively. This indicates that \textbf{CA-TTS} more effectively leverages additional samples, with the performance gap widening as sample count increases from 1 to 32. While all methods start at similar accuracy with a single sample ($\sim$25-30\%), our approach scales to over 45\% accuracy, significantly outperforming the baselines' plateau at $\sim$35\%. This robust scaling confirms that calibrated confidence enables more efficient test-time computation. Additional scaling results are provided in Appendix~\ref{app:c}.

\begin{figure*}[h]
    \centering
    \includegraphics[width=0.95\linewidth]{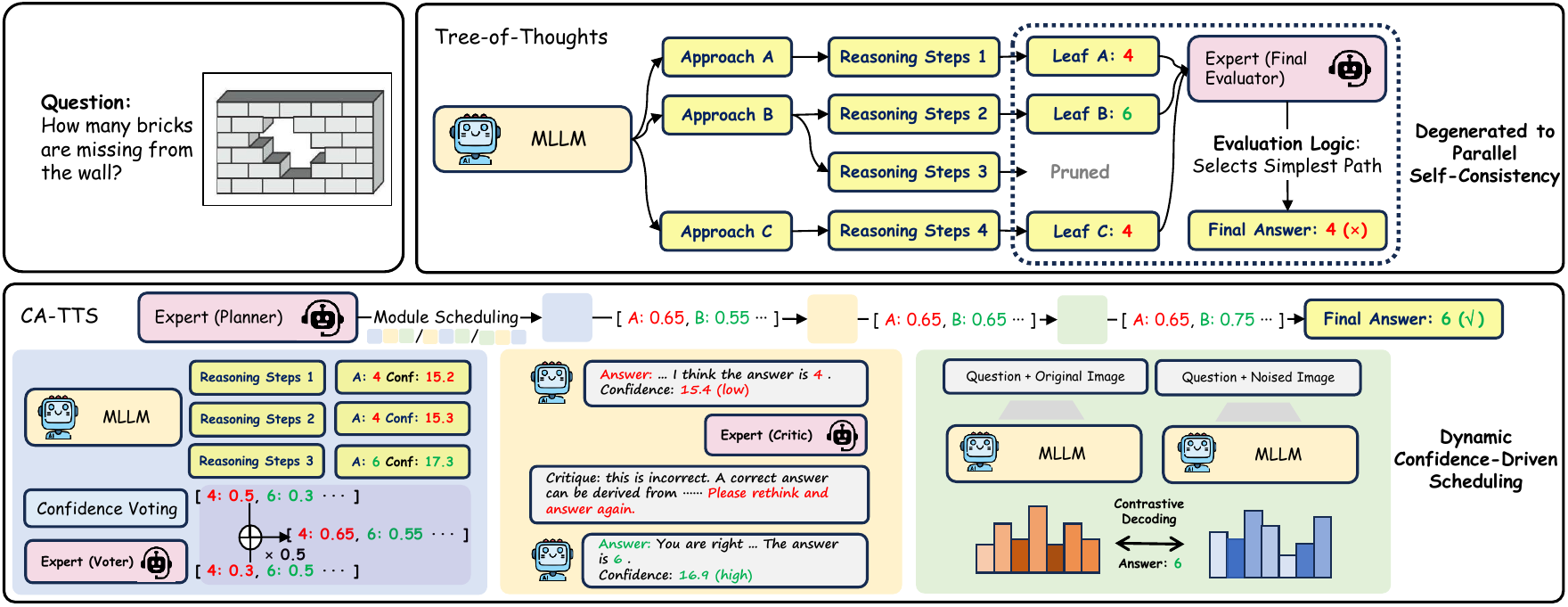}
    \caption{\textbf{A case study comparing the reasoning processes of ToT \cite{yao2023treethoughtsdeliberateproblem} and CA-TTS (Ours).} ToT (upper) conducts a complex tree search that remains vulnerable to a single-point-of-failure in its final evaluation, leading it to the incorrect answer. In contrast, Our method (bottom) demonstrates a multi-stage, resilient process: an initial error from \textbf{Self-Consistency} (Answer: 4) is corrected by \textbf{Self-Reflection} (Answer: 6) and confirmed by \textbf{Self-Check}.}
    \label{fig:case_study}
\end{figure*}

\noindent \textbf{d) CDRL enables knowing when visual evidence is insufficient.} As shown in Table~\ref{tab:table4}, \textbf{CDRL} training significantly enhances the model's perceptual sensitivity across multiple visual uncertainty conditions. The baseline model (Qwen2.5-VL-7B-Instruct) exhibits minimal confidence drops (CD) when visual input is compromised—near-zero or even positive CD values for Occlusion ($-0.24$), Viewpoint ($+0.09$), and Mosaic ($+0.11$) conditions indicate poor awareness of visual degradation. After \textbf{CDRL} training, the model demonstrates substantially larger confidence drops across all perturbation types: Noised ($-1.39$), Occlusion ($-1.13$), Viewpoint ($-1.29$), and Mosaic ($-0.86$). This represents a 4-8$\times$ enhancement in visual sensitivity on average. Additionally, \textbf{CDRL} improves calibration metrics (ECE) and uncertainty quantification (AUC) across all conditions, demonstrating that the model learns to properly assess perception quality and reduce confidence when visual evidence is insufficient.


\begin{table}[tbp]
\centering
\caption{\textbf{Analysis of model confidence sensitivity under different visual uncertainty conditions on Math-Vision.} We compare the base model with our \textbf{CDRL}-trained model across five visual conditions. \textbf{CD} (Confidence Drop) measures sensitivity to visual perturbations relative to the original image . \textbf{ECE} and \textbf{AUC} measure calibration quality. Best results are in \textbf{bold}.}
\label{tab:table4}
\resizebox{\columnwidth}{!}{
\begin{tabular}{l|ccc|ccc}
\toprule
 & \multicolumn{3}{c|}{Qwen2.5-VL-7B-Instruct} & \multicolumn{3}{c}{\textbf{+CDRL (Ours)}} \\
Visual Uncertainty & CD$\downarrow$ & ECE$\downarrow$ & AUC$\uparrow$ & CD$\downarrow$ & ECE$\downarrow$ & AUC$\uparrow$ \\
\midrule
Origin & 0 & 64.57 & 54.81 & 0 & \textbf{62.24} & \textbf{59.42} \\
Noised & -0.32 & 66.19 & 60.00 & \textbf{-1.39} & \textbf{63.75} & \textbf{60.19} \\
Occlusion & -0.24 & 66.19 & 53.88 & \textbf{-1.13} & \textbf{65.05} & \textbf{56.34} \\
Viewpoint & +0.09 & 64.99 & 55.19 & \textbf{-1.29} & \textbf{64.18} & \textbf{57.26} \\
Mosaic & +0.11 & 65.41 & 60.01 & \textbf{-0.86} & \textbf{63.07} & \textbf{61.68} \\
\bottomrule
\end{tabular}
}
\end{table}


\subsection{Case Study: CA-TTS vs. ToT}
\label{case_study}
As shown in Figure~\ref{fig:case_study}, Our method highlights a robust, decoupled reasoning process. As illustrated, the Expert (Planner) first schedules modules. In the \textit{Self-Consistency} phase, the model may initially converge on an incorrect answer (e.g., 4), even with voter intervention. However, the process continues: in the \textit{Self-Reflection} phase, the expert acts as a \textit{Critique}, providing a new incentive signal that guides the model to correct its answer to 6. This is subsequently solidified in the \textit{Self-Check} phase using visual-contrastive decoding. This demonstrates our method's key advantages: it is robust, features multiple decoupled verification stages, and benefits from continuous and varied incentive signals.

In contrast, tree-based methods like ToT \cite{yao2023treethoughtsdeliberateproblem}, while exploratory, are often more cumbersome and possess a critical vulnerability: a heavy reliance on the performance of a single-pass, final evaluation model. When all paths reach their leaf nodes, this single evaluation determines the outcome. As our comparative example illustrates, a flaw in this single evaluation can cause the entire complex exploration to converge on the wrong answer (i.e., 4). Our approach avoids this single point of failure. As discussed in Section \ref{sec:self}, our method shows significant gains even with self-evaluation, proving it is less sensitive to the expert model's perfection. Thus, \textbf{CA-TTS} offers a more resilient and efficient process through multi-stage validation, unlike ToT's high-stakes dependency on one-shot evaluation. 

\section{Conclusion}

This work identifies \textit{perceptual bluntness} as a root cause of hallucination in MLLMs: a model that reasons before perceiving will inevitably produce unreliable answers. We demonstrate that a synergistic \textit{Perceive-then-Reason} approach is essential. The \textbf{CDRL} training stage first instills perceptual calibration, creating the necessary foundation that enables our adaptive \textbf{CA-TTS} framework to successfully orchestrate its reasoning modules. Extensive experiments across four challenging benchmarks validate our approach. This finding signals a paradigm shift for MLLM research: future efforts must move beyond text-level preference tuning and co-optimize visual grounding with confidence calibration to build truly robust, self-aware systems that know what they see and when they don't know.


\section{Acknowledgments}

This work was supported by the National Natural Science Foundation of China under Grant 42394060 and 42394064, the Earth System Big Data Platform of the School of Earth Sciences, Zhejiang University, and the Zhejiang University - Jolly Pharmaceutical Joint R\&D Center for Intelligent Empowerment in Food and Medicine.

\label{sec:result}

%% file: sec/X_suppl.tex
\clearpage
\setcounter{page}{1}
\maketitlesupplementary

\begin{figure*}[ht]
    \centering
    \includegraphics[width=0.95\linewidth]{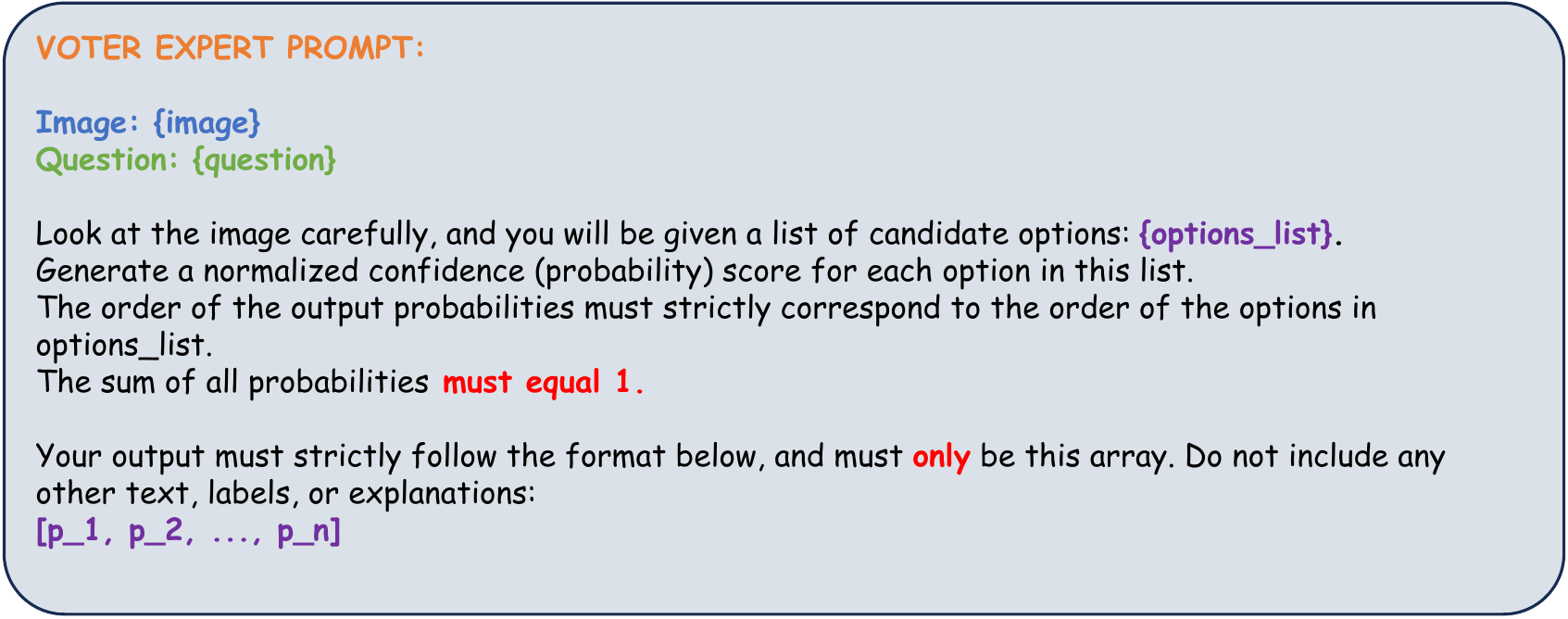}
    \caption{The prompt template used for the Voter Expert. The model acts as a discriminator to assign probability scores to candidate choices, facilitating confidence-weighted voting.}
    \label{fig:p-voter}
\end{figure*}






\renewcommand{\thesection}{\Alph{section}} 
\renewcommand{\thesubsection}{\thesection.\arabic{subsection}} 

\setcounter{section}{0} 


\section{Prompt Templates} 
\label{app:a}

In this section, we provide the specific prompt templates used within our proposed framework. As illustrated in Figure~2, our \textit{Confidence-Aware Test-Time Scaling} (CA-TTS) system employs an Expert Model to fulfill specific roles (specifically as a \textbf{Voter} and a \textbf{Critic}) to coordinate the self-consistency and self-reflection modules.

\subsection{Voter Expert Prompt}
The Voter Expert is utilized during the confidence voting process of the \textit{Self-Consistency} phase. It is tasked with analyzing the image and question to assign normalized confidence probabilities to a provided list of candidate options. The specific prompt structure is shown in Figure~\ref{fig:p-voter}.

\subsection{Critic Expert Prompt}
\begin{figure*}[ht]
    \centering
    \includegraphics[width=0.95\linewidth]{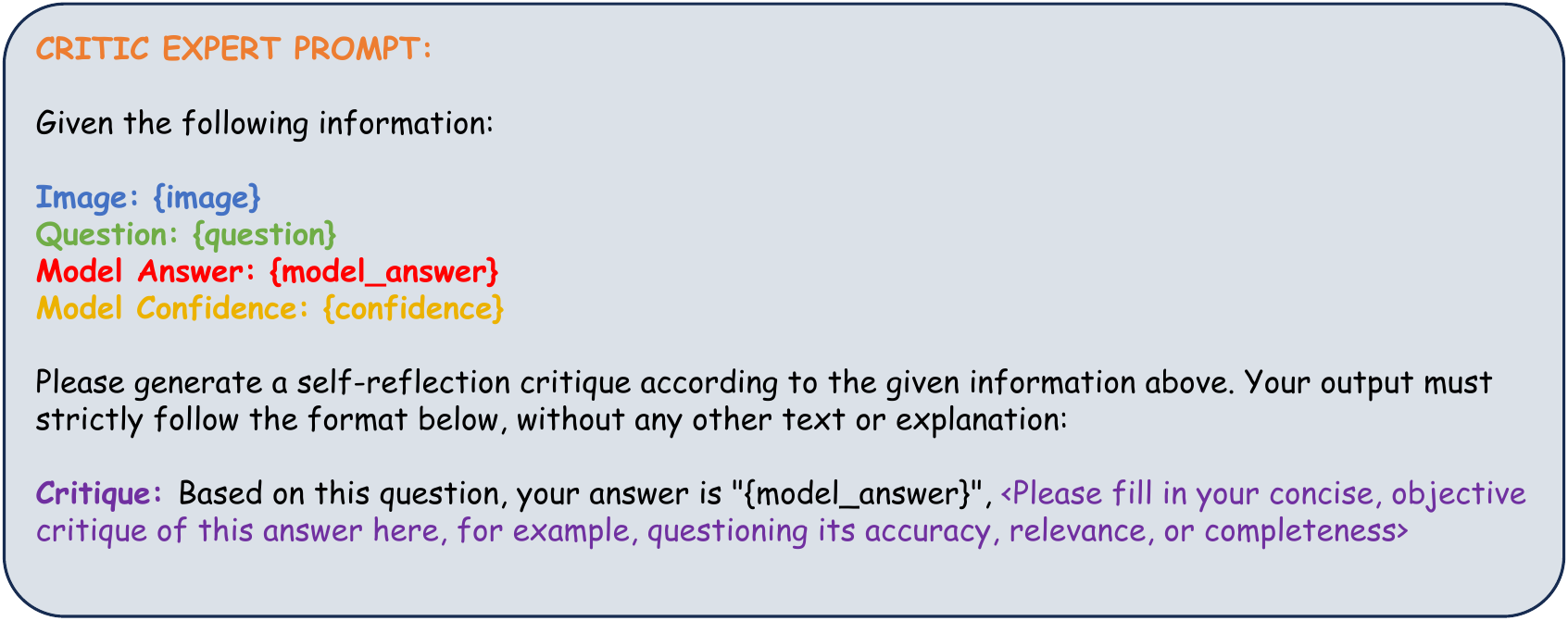}
    \caption{The prompt template used for the Critic Expert. This prompt induces the expert model to generate a critique based on the original input and the model's initial low-confidence response, aiding in the self-reflection process.}
    \label{fig:p-critique}
\end{figure*}

The Critic Expert is engaged during the \textit{Self-Reflection} phase. It evaluates the model's initial answer and confidence level to provide a constructive critique. This critique guides the Multi-Modal Large Language Model (MLLM) to rethink and refine its answer. The prompt structure is detailed in Figure~\ref{fig:p-critique}.

\begin{table*}[h]
\centering
\caption{\textbf{Performance on Qwen3-VL-2B-Thinking.} Comparison of CA-TTS against other TTS baselines. Abbreviations follow previous tables. Best results are \textbf{bold}, and second-best results are \underline{underlined}. The ``Overall Avg'' represents the mean of the ``ALL'' scores across the four datasets.}
\resizebox{\linewidth}{!}{
\begin{tabular}{l ccc ccc ccc ccc c}
\toprule
& \multicolumn{6}{c}{\textbf{Math Reasoning}} & \multicolumn{6}{c}{\textbf{General VQA}} & \\
\cmidrule(lr){2-7} \cmidrule(lr){8-13}
\textbf{Methods} & \multicolumn{3}{c}{\textbf{Math-Vista$_{testmini}$}} & \multicolumn{3}{c}{\textbf{Math-Vision$_{testmini}$}} & \multicolumn{3}{c}{\textbf{MMStar}} & \multicolumn{3}{c}{\textbf{MMMU}} & \textbf{Overall} \\
\cmidrule(lr){2-4} \cmidrule(lr){5-7} \cmidrule(lr){8-10} \cmidrule(lr){11-13} \cmidrule(lr){14-14}
(Qwen3-VL-2B-Think) & OE & MC & ALL & OE & MC & ALL & PE & RE & ALL & STEM & HASS & ALL & \textbf{Avg.} \\
\midrule
Pass@1 & 64.95 & 82.51 & 73.60 & 47.69 & 50.00 & 48.82 & 62.79 & 56.00 & 61.24 & 58.84 & 66.95 & 61.40 & 61.27 \\
Majority Voting & 72.30 & \underline{89.86} & \underline{80.95} & \textbf{60.87} & \underline{62.50} & \underline{61.70} & 67.10 & 66.49 & \underline{66.24} & 72.90 & 81.01 & 75.46 & \underline{71.09} \\
DeepConf & \underline{72.77} & 87.92 & 80.24 & \underline{52.17} & \underline{62.50} & 57.45 & \underline{67.36} & \underline{68.09} & \underline{66.24} & \underline{73.83} & \underline{82.28} & \underline{75.93} & 69.97 \\
\textbf{Ours (CA-TTS)} & \textbf{77.46} & \textbf{90.34} & \textbf{83.81} & \textbf{60.87} & \textbf{66.67} & \textbf{63.83} & \textbf{69.71} & \textbf{75.53} & \textbf{70.36} & \textbf{78.50} & \textbf{83.54} & \textbf{79.63} & \textbf{74.41} \\
\bottomrule
\end{tabular}
}
\label{tab:qwen3_performance}
\end{table*}

\begin{table*}[h]
\centering
\caption{\textbf{Detailed Scaling Results.} Comparison of performance with varying number of samples ($N$). Best results are \textbf{bold}.}
\resizebox{0.9\linewidth}{!}{
\begin{tabular}{ll cccccc}
\toprule
& & \multicolumn{6}{c}{\textbf{Number of Samples ($N$)}} \\
\cmidrule(lr){3-8}
\textbf{Dataset} & \textbf{Method} & \textbf{1} & \textbf{2} & \textbf{4} & \textbf{8} & \textbf{16} & \textbf{32} \\
\midrule

 & Majority Voting & 26.38 & 25.00 & 29.93 & 30.26 & 30.92 & 34.41 \\
\textbf{Math-Vision}$_{testmini}$ & DeepConf & 26.38 & 26.64 & 28.95 & 28.95 & 30.26 & 32.15 \\
 & \textbf{Ours (CDRL+CA-TTS)} & \textbf{30.60} & \textbf{34.21} & \textbf{41.78} & \textbf{43.75} & \textbf{46.38} & \textbf{48.44} \\
\midrule

 & Majority Voting & \textbf{64.70} & 67.78 & 68.56 & 68.39 & 73.11 & 75.60 \\
\textbf{Math-Vista}$_{testmini}$ & DeepConf & \textbf{64.70} & 67.89 & 67.69 & 69.28 & 71.32 & 75.50 \\
 & \textbf{Ours (CDRL+CA-TTS)} & 64.15 & \textbf{72.15} & \textbf{73.80} & \textbf{74.19} & \textbf{77.85} & \textbf{80.60} \\
\midrule

 & Majority Voting & 60.20 & 60.55 & 63.68 & 69.00 & 64.56 & 69.96 \\
\textbf{MMStar} & DeepConf & 60.20 & 59.63 & 62.88 & 61.87 & 64.08 & 69.77 \\
 & \textbf{Ours (CDRL+CA-TTS)} & \textbf{61.21} & \textbf{63.30} & \textbf{71.16} & \textbf{71.27} & \textbf{70.87} & \textbf{74.03} \\
\midrule

 & Majority Voting & 48.77 & 54.14 & 55.84 & 57.18 & 58.87 & 58.62 \\
\textbf{MMMU} & DeepConf & 48.77 & 54.61 & 56.31 & 56.24 & 58.25 & 58.00 \\
 & \textbf{Ours (CDRL+CA-TTS)} & \textbf{52.63} & \textbf{61.96} & \textbf{65.65} & \textbf{66.28} & \textbf{68.72} & \textbf{69.58} \\
\bottomrule
\end{tabular}
}
\label{tab:scaling_results}
\end{table*}

\begin{table*}[htbp]
\centering
\caption{\textbf{Performance of Different Expert Models.} Abbreviations: OE (Open-Ended), MC (Multi-Choice), PE (Perception), RE (Reasoning), STEM (Science, Technology, Engineering, and Mathematics), HASS (Humanities, Arts, and Social Sciences), and ALL (Overall). Best results are \textbf{bold}, and second-best results are \underline{underlined}. The ``Overall Avg'' column is calculated as the average of the ``ALL'' scores from the four datasets.}
\resizebox{\linewidth}{!}{
\begin{tabular}{l ccc ccc ccc ccc c}
\hline
& \multicolumn{6}{c}{\textbf{Math Reasoning}} & \multicolumn{6}{c}{\textbf{General VQA}} & \\
\cmidrule(lr){2-7} \cmidrule(lr){8-13}
\textbf{Models/Datasets} & \multicolumn{3}{c}{\textbf{Math-Vista$_{testmini}$}} & \multicolumn{3}{c}{\textbf{Math-Vision$_{testmini}$}} & \multicolumn{3}{c}{\textbf{MMStar}} & \multicolumn{3}{c}{\textbf{MMMU}} & \textbf{Overall} \\
\cmidrule(lr){2-4} \cmidrule(lr){5-7} \cmidrule(lr){8-10} \cmidrule(lr){11-13} \cmidrule(lr){14-14}
& OE & MC & ALL & OE & MC & ALL & PE & RE & ALL & STEM & HASS & ALL & \textbf{Avg} \\
\hline
Majority Voting & 68.39 & 73.73 & 69.80 & 26.20 & 33.24 & 30.08 & 60.20 & 71.20 & 64.00 & 53.20 & 62.50 & 57.53 & 55.35 \\
Qwen2.5-VL-7B & 69.49 & 77.99 & 73.00 & 30.91 & 34.03 & 32.57 & \underline{65.33} & 71.47 & 64.59 & 53.35 & 65.99 & 58.46 & 57.16 \\
Qwen2.5-VL-72B & 67.90 & 78.62 & 74.20 & 28.52 & 37.70 & 34.21 & 61.00 & 71.60 & 64.33 & 55.08 & 61.56 & 58.11 & 57.71 \\
Qwen-VL-Max & 69.35 & 79.14 & 74.47 & 35.71 & 36.13 & 35.97 & 61.20 & 73.40 & 64.80 & 56.25 & 67.05 & 61.45 & 59.17 \\
GPT-5 & \underline{73.74} & \underline{82.71} & \underline{78.20} & \textbf{38.94} & \underline{43.93} & \underline{41.45} & 64.00 & \underline{77.00} & \underline{70.40} & \textbf{61.85} & \underline{72.62} & \textbf{66.51} & \underline{64.14} \\
Gemini-2.5-Pro & \textbf{74.19} & \textbf{84.67} & \textbf{79.50} & \underline{38.44} & \textbf{45.60} & \textbf{42.35} & \textbf{67.80} & \textbf{77.20} & \textbf{71.27} & \underline{59.90} & \textbf{73.50} & \underline{66.28} & \textbf{64.85} \\
\hline
\end{tabular}
}
\label{tab:expert_performance}
\end{table*}

\section{Training Datasets Details}
\label{app:b}

Previous research \cite{muennighoff2025s1simpletesttimescaling} has shown that efficient training driven by a small amount of high-quality data can significantly elicit the scaling ability of models. Inspired by this, we aim to construct a refined and efficient Reinforcement Learning (RL) training set, denoted as $D_{\text{RL}}$.

To establish the source data pool $D_{\text{source}}$, we aggregated data from six public benchmarks, comprising three mathematical reasoning datasets and three general-purpose VQA datasets. These benchmarks include:
\begin{itemize}
    \item \textbf{MathVerse} \cite{zhang2024mathversedoesmultimodalllm}: A rigorous visual mathematical benchmark designed to decouple visual and textual dependencies, evaluating true multimodal reasoning capabilities across diverse diagrams.
    \item \textbf{DynaMath} \cite{zou2025dynamathdynamicvisualbenchmark}: A dynamic visual benchmark that assesses model generalization in mathematical reasoning by programmatically generating diverse problem variations.
    \item \textbf{LogicVista} \cite{xiao2024logicvistamultimodalllmlogical}: A comprehensive benchmark focused on logical reasoning in multimodal contexts, spanning tasks from puzzle solving to complex diagrammatic analysis.
    \item \textbf{RealWorldQA} \cite{xai_realworldqa}: A benchmark evaluating real-world spatial understanding and physical reasoning capabilities, primarily derived from challenging real-world environments.
    \item \textbf{MMBench} \cite{liu2024mmbenchmultimodalmodelallaround}: An all-around multimodal evaluation pipeline that utilizes a circular evaluation strategy to robustly assess perception and reasoning across diverse tasks.
    \item \textbf{MMMU-Pro} \cite{yue2025mmmuprorobustmultidisciplinemultimodal}: A robust and refined version of the massive multidisciplinary benchmark MMMU, specifically designed to strictly evaluate expert-level reasoning by filtering out text-shortcut samples.
\end{itemize}

The aggregated source pool $D_{\text{source}}$ contains a total of $N = 11764$ data points. To construct the final $D_{\text{RL}}$, we implemented a two-phase process involving strict data filtering followed by visual augmentation.

\noindent \textbf{Phase 1: Data Filtering.} 
To ensure high training efficiency, we defined a set of strict manual filtering criteria, $\mathcal{C}$, which comprehensively evaluates each data point $d$ based on its \textbf{Quality}, \textbf{Difficulty}, and \textbf{Diversity}. We apply these criteria to filter out a high-quality intermediate set $D_{\text{Filtered}}$:
\begin{equation}
D_{\text{Filtered}} = \{ d \in D_{\text{source}} \mid \text{Evaluate}(d, \mathcal{C}) = \text{True} \}
\end{equation}
Here, $\text{Evaluate}(d, \mathcal{C})$ represents the manual assessment process. Through this procedure, we selected $M=1936$ high-quality data points from the source pool.

\noindent \textbf{Phase 2: Data Augmentation and Image-Pair Construction.} 
Subsequently, to enhance the model's robustness and perception of critical visual regions, we applied specific visual augmentations to $D_{\text{Filtered}}$. We utilize intermediate-layer attention maps, $\text{Attn}_{\text{CLIP}}$, from the CLIP \cite{radford2021learningtransferablevisualmodels} visual encoder to identify core semantic regions within an image $i$. We define a noise-injection function $\mathcal{G}_{\text{noise}}$ that applies noise to the original image based on its attention map, yielding a perturbed image $i' = \mathcal{G}_{\text{noise}}(i, \text{Attn}_{\text{CLIP}}(i))$.

The final RL training set $D_{\text{RL}}$ expands each filtered data point $d_f = (i, q, a)$ into a tuple containing an \textit{[Original Image, Noised Image]} pair. The formal definition is as follows:
\begin{equation}
\begin{split}
    D_{\text{RL}} = \{ ((i, i'), q, a) \mid (i, q, a) \in D_{\text{Filtered}}, 
    \\ \text{ where } i' = \mathcal{G}_{\text{noise}}(i, \text{Attn}_{\text{CLIP}}(i)) \}
\end{split}
\end{equation}
This resulting dataset $D_{\text{RL}}$, containing $M=1936$ data pairs, is employed in our subsequent RL training to teach the model to robustly locate and reason about key information even in the presence of visual disturbances.

\section{Additional Results} 
\label{app:c}
\subsection{Tables} 
\noindent\textbf{a) Generalizability on Emerging Reasoning-Capable MLLMs.} To further scrutinize the universality and robustness of our approach, we extended our evaluation to \textit{Qwen3-VL-2B-Thinking}, a representative of the latest MLLMs equipped with intrinsic chain-of-thought (CoT) capabilities. This experiment investigates a pivotal question: does CA-TTS provide additive value to models that already possess optimized ``thinking" processes?

As detailed in Table~\ref{tab:qwen3_performance}, the answer is affirmative. CA-TTS consistently outperforms both the standard Majority Voting and the DeepConf baseline across all evaluated benchmarks. Notably, our framework achieves a commanding \textbf{Overall Avg of 74.41}, securing a clear lead over the Majority Voting baseline (71.09). 

The performance gains are particularly pronounced in reasoning-intensive domains. On \textit{Math-Vista}, CA-TTS elevates the score to \textbf{83.81}, significantly surpassing the Pass@1 baseline of 73.60. Moreover, on the comprehensive \textit{MMMU} benchmark, our method achieves a remarkable \textbf{79.63} (vs. 75.46 for Majority Voting). These results suggest that even for models with native CoT designs, CA-TTS effectively modulates the reasoning trajectory to correct errors and refine outputs, thereby unlocking a higher ceiling of performance in both mathematical reasoning and general visual understanding.

\label{app:c-b}
\noindent \textbf{b) Additional Model Scaling Results.} While the main text visualizes the scaling trends primarily on the \textit{Math-Vision} dataset due to space constraints, here we present the comprehensive numerical data across all four benchmarks in Table~\ref{tab:scaling_results}. This detailed breakdown allows for a granular analysis of how performance evolves with the number of samples ($N$) ranging from 1 to 32.

The results clearly demonstrate the superior scaling efficiency of \textbf{Our Method (CDRL+CA-TTS)}. Compared to standard Majority Voting and DeepConf, our approach consistently achieves higher accuracy gains as the sample size increases. Notably, on the \textit{MMMU} and \textit{Math-Vision} datasets, our method maintains a substantial lead at every scaling step. For instance, at $N=32$, CDRL+CA-TTS outperforms the best baseline on Math-Vision by a margin of over 14\% (48.44\% vs 34.41\%), validating that our confidence-aware test-time scaling strategy effectively leverages increased test-time computation to resolve complex visual reasoning tasks.

\noindent \textbf{c) Additional Results of Performance by Using Different Expert Models.} Quantitative results in Table~\ref{tab:expert_performance} further substantiate these observations. The framework exhibits a positive correlation between the capability of the expert model and the final performance. Specifically, Gemini-2.5-Pro achieves the state-of-the-art performance with an \textbf{Overall Avg of 64.85}, significantly outperforming the Majority Voting baseline (55.35). It dominates across most benchmarks, particularly in \textit{Math-Vista} (79.50) and \textit{MMStar} (71.27). GPT-5 follows closely as the second-best performer with an Overall Avg of 64.14, while demonstrating superior capability in the \textit{MMMU} benchmark (66.51 compared to Gemini's 66.28), specifically in STEM tasks. Notably, even smaller models like Qwen2.5-VL-7B provide a clear boost over the baseline (57.16 vs. 55.35), validating the effectiveness of our framework regardless of the expert model's size. This trend confirms that our approach effectively leverages the distinct strengths of various foundation models, from reasoning-heavy tasks in Math Reasoning datasets to general visual question answering.

\noindent \textbf{d) Compatibility with Existing Frameworks.} To validate whether our proposed CA-TTS can serve as a universal plug-in to enhance existing models, we integrated it with several state-of-the-art baselines, utilizing Gemini-2.5-Pro as the expert model. As shown in Table~\ref{tab:framework_compatibility}, under identical settings, CDRL combined with CA-TTS achieves the best overall average performance (64.9\%), outperforming the second-best framework (We-Think) by a clear margin of 3.9\%. This demonstrates that calibrated confidence provides a fundamentally stronger foundation for effective test-time scaling.

\noindent \textbf{e) Ablation Study and Inference Cost.} We conducted a comprehensive ablation study to isolate the contribution of each module within the CA-TTS framework and analyzed their corresponding inference time costs. As detailed in Table~\ref{tab:ablation_cost}, the removal of any module leads to a noticeable performance drop, validating our carefully coordinated design. Regarding inference efficiency, compared to the standard Majority Voting baseline, CA-TTS consumes only 0.91$\times$ more time while delivering a substantial 8.4\% increase in average accuracy, proving it is highly efficient within the paradigm of test-time scaling.

\begin{table*}
\centering
\caption{\textbf{Baseline Models Results with Gemini 2.5 Pro.} Comparison of various frameworks when equipped with our CA-TTS plug-in.}
\resizebox{0.75\linewidth}{!}{%
\begin{tabular}{l ccccc}
\toprule
\textbf{Framework + CA-TTS} & \textbf{Math-Vista} & \textbf{Math-Vision} & \textbf{MMStar} & \textbf{MMMU} & \textbf{Average} \\
\midrule
Qwen2.5-VL-7B & 75.8 & 39.1 & 65.6 & 59.2 & 59.9 \\
R1-OneVision & 70.9 & 38.2 & 60.9 & 59.8 & 57.5 \\
VL-Rethinker & 74.3 & 38.1 & 64.3 & 57.1 & 58.5 \\
We-Think & 77.1 & 39.8 & 66.8 & 60.2 & 61.0 \\
\textbf{Ours (CDRL)} & \textbf{79.5} & \textbf{42.4} & \textbf{71.3} & \textbf{66.3} & \textbf{64.9} \\
\bottomrule
\end{tabular}%
}
\label{tab:framework_compatibility}
\end{table*}

\begin{table*}
\centering
\caption{\textbf{Ablation Study and Inference Cost.} Evaluation of individual modules and their corresponding time consumption.}
\resizebox{0.85\linewidth}{!}{%
\begin{tabular}{l cccccc} 
\toprule
\textbf{Settings} & \textbf{Math-Vista} & \textbf{Math-Vision} & \textbf{MMStar} & \textbf{MMMU} & \textbf{Avg.} & \textbf{Time Cost (s)} \\ 
\midrule
\textbf{Ours (CA-TTS)} & \textbf{79.5} & \textbf{42.4} & \textbf{71.3} & \textbf{66.3} & \textbf{64.9} & 11.03 \\
w/o Self-Consistency & 70.5 & 35.0 & 67.4 & 58.4 & 57.8 & 4.66\\
w/o Self-Reflection & 74.6 & 37.9 & 69.1 & 65.5 & 61.8 & 8.55\\
w/o Self-Check & 74.2 & 39.1 & 70.5 & 65.9 & 62.4 & 8.85\\
Majority Voting & 69.8 & 30.1 & 69.0 & 57.5 & 56.6 & 5.76\\
\bottomrule
\end{tabular}%
}
\label{tab:ablation_cost}
\end{table*}

\subsection{Visualization} 

\begin{figure}
    \centering
    \includegraphics[width=0.8\linewidth]{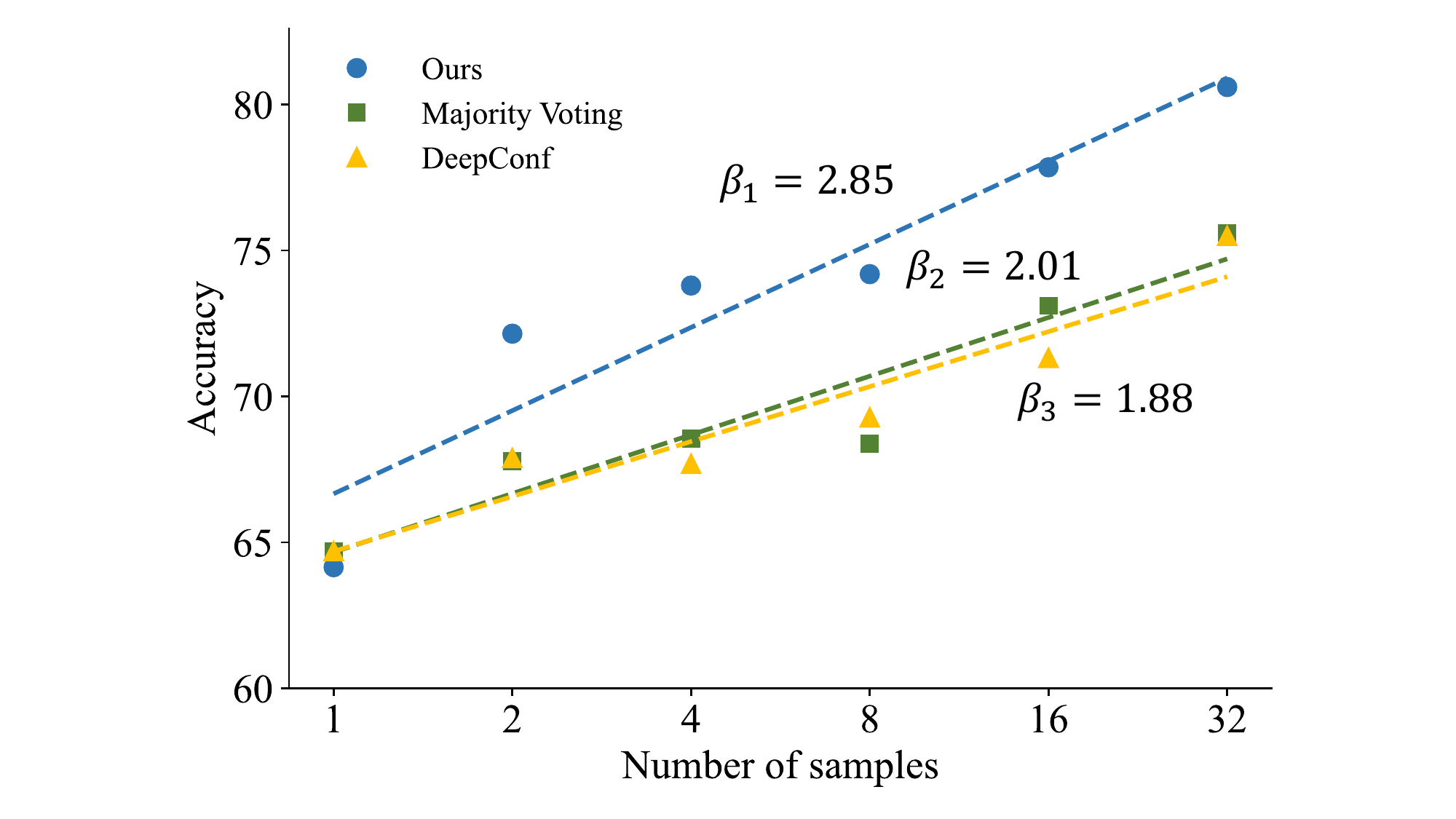}
    \caption{\textbf{Scaling Superiority on Math-Vista.} This plot compares the accuracy of Our method (blue dots) against the Majority Voting (green squares) and DeepConf (yellow triangles) baselines as the number of samples increases from 1 to 32. The results show that the slope of our trendline ($\beta_1 = 2.85$) is substantially steeper than those of the baseline methods ($\beta_2 = 2.01$ and $\beta_3 = 1.88$), indicating that the performance advantage and potential of our method widens as more samples are provided.}
    \label{fig:fig7}
\end{figure}

\begin{figure}
    \centering
    \includegraphics[width=0.8\linewidth]{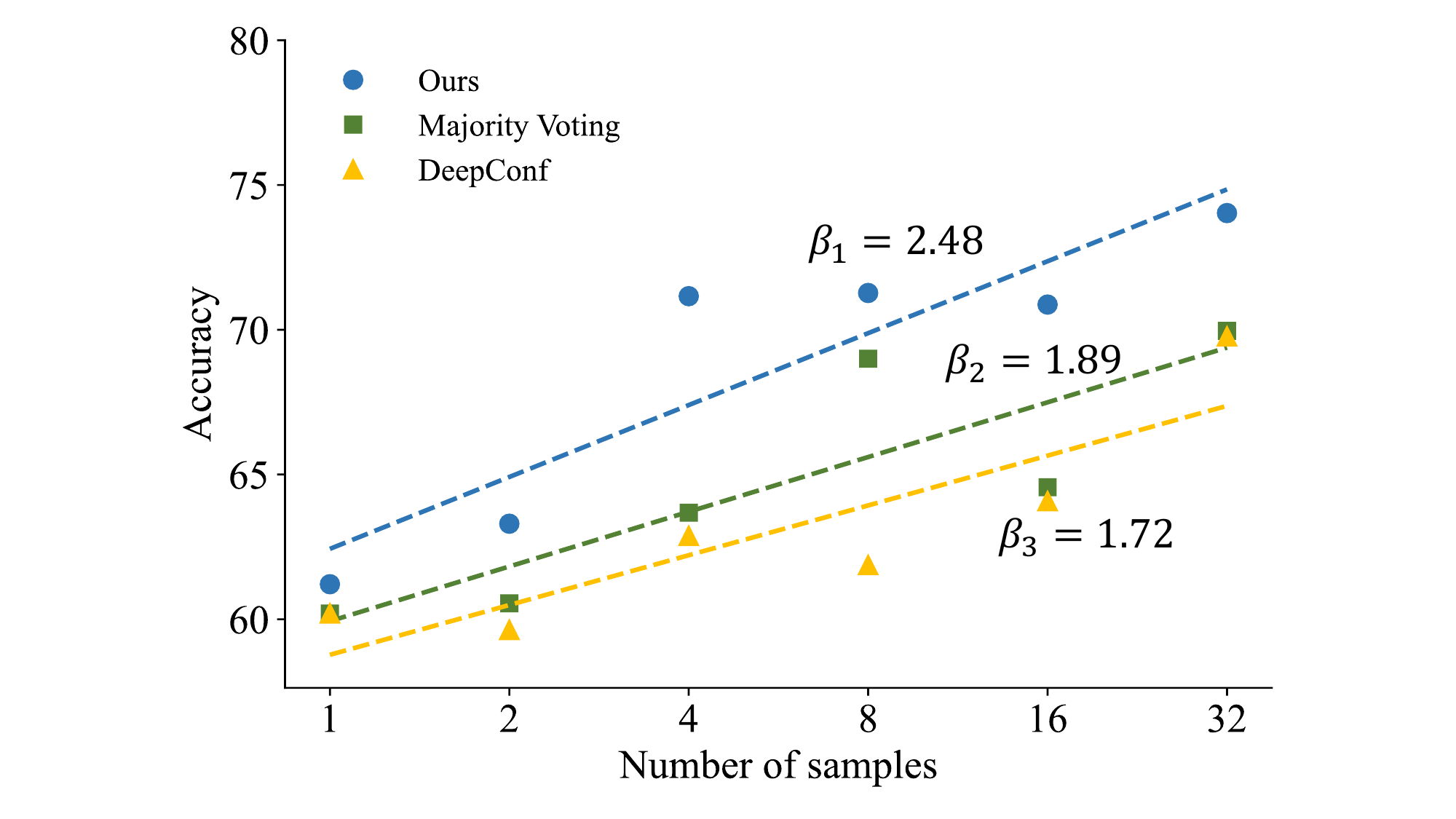}
    \caption{\textbf{Scaling Superiority on MMStar.} This plot compares the accuracy of Our method (blue dots) against the Majority Voting (green squares) and DeepConf (yellow triangles) baselines as the number of samples increases from 1 to 32. The results show that the slope of our trendline ($\beta_1 = 2.48$) is substantially steeper than those of the baseline methods ($\beta_2 = 1.89$ and $\beta_3 = 1.72$), indicating that the performance advantage and potential of our method widens as more samples are provided.}
    \label{fig:fig8}
\end{figure}

\begin{figure}
    \centering
    \includegraphics[width=0.8\linewidth]{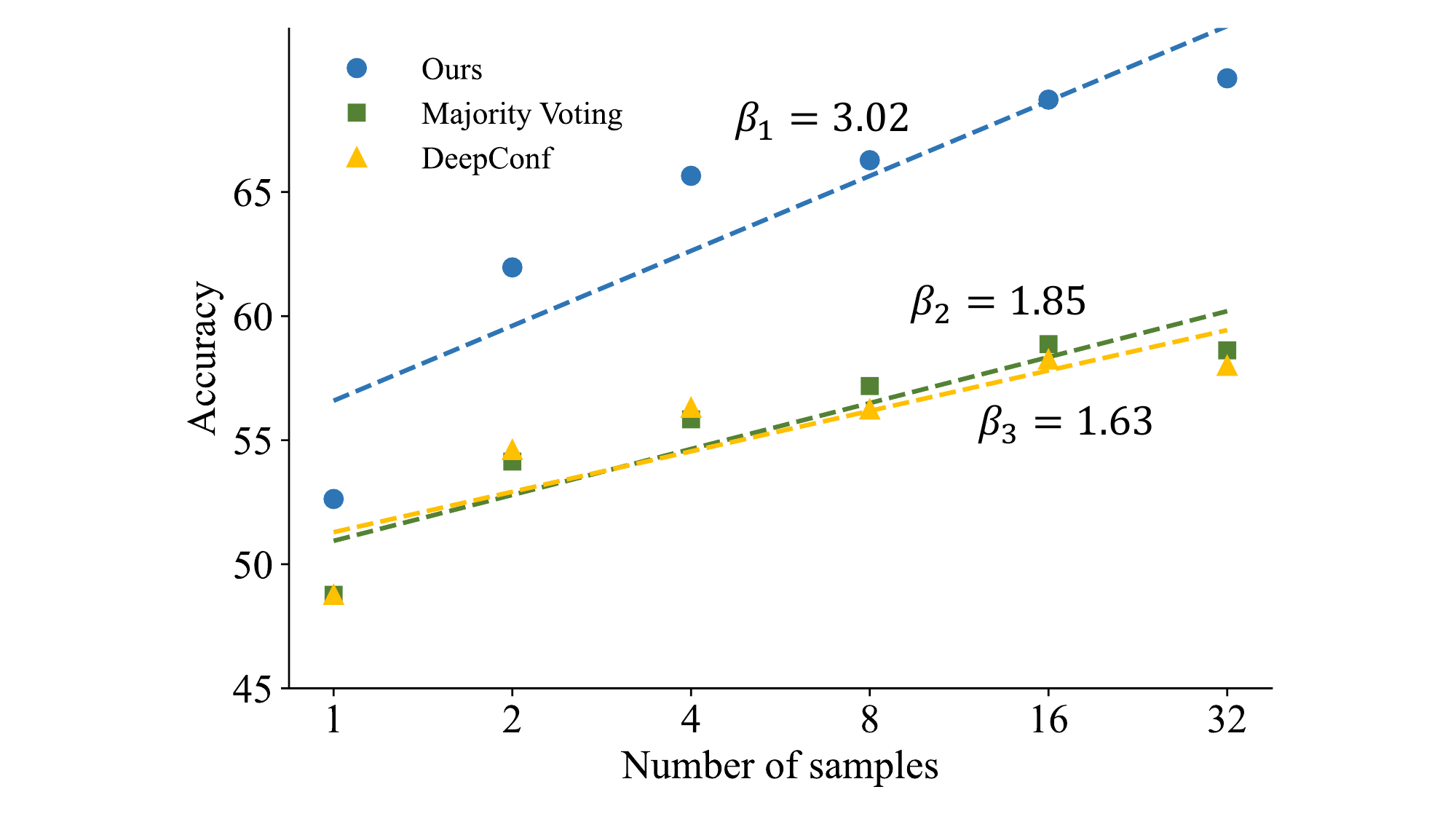}
    \caption{\textbf{Scaling Superiority on MMMU.} This plot compares the accuracy of Our method (blue dots) against the Majority Voting (green squares) and DeepConf (yellow triangles) baselines as the number of samples increases from 1 to 32. The results show that the slope of our trendline ($\beta_1 = 3.02$) is substantially steeper than those of the baseline methods ($\beta_2 = 1.85$ and $\beta_3 = 1.63$), indicating that the performance advantage and potential of our method widens as more samples are provided.}
    \label{fig:fig9}
\end{figure}

\begin{figure}
  \centering
  \includegraphics[width=1.0\linewidth]{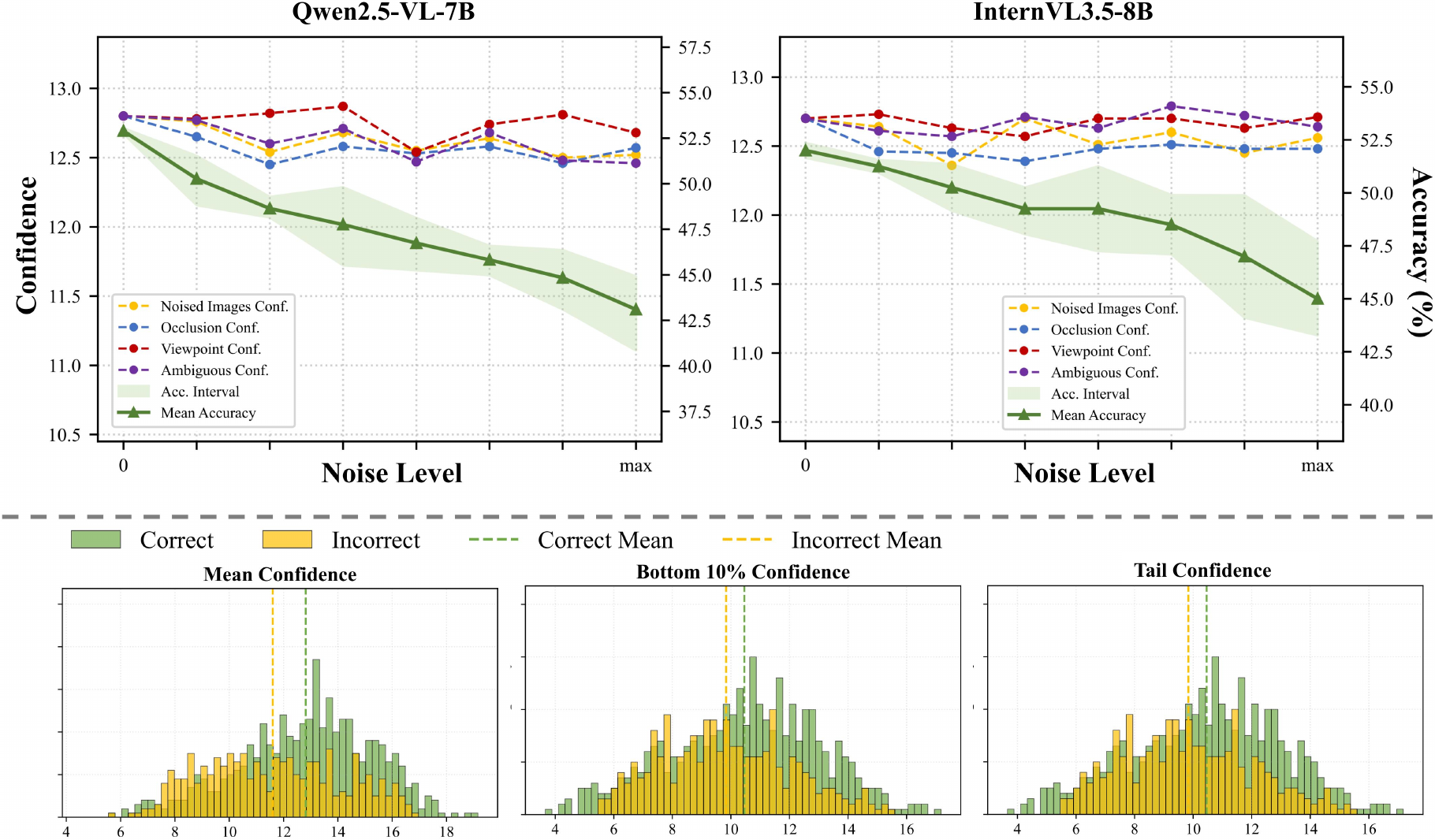}
  \caption{\textbf{Generality of Confidence Miscalibration under Visual Degradation.} We tested the impact of various visual degradation types on two representative MLLMs. The overlapping confidence distributions indicate that MLLMs struggle to calibrate their confidence correctly when handling degraded visual inputs, highlighting the necessity of our Confidence-Aware framework.}
  \label{fig:rebuttal_dist}
\end{figure}

As shown in Figure~\ref{fig:fig7},~\ref{fig:fig8} and~\ref{fig:fig9}, we illustrates additional visualized scaling results. Detailed analysis can be found in figure captions, Section \ref{4:scaling} and Appendix~\ref{app:c-b}. Additionally, Figure~\ref{fig:rebuttal_dist} illustrates the generality of confidence miscalibration under visual degradation.

\subsection{More Case Studies} 

\begin{figure*}[h]
    \centering
    \includegraphics[width=\linewidth]{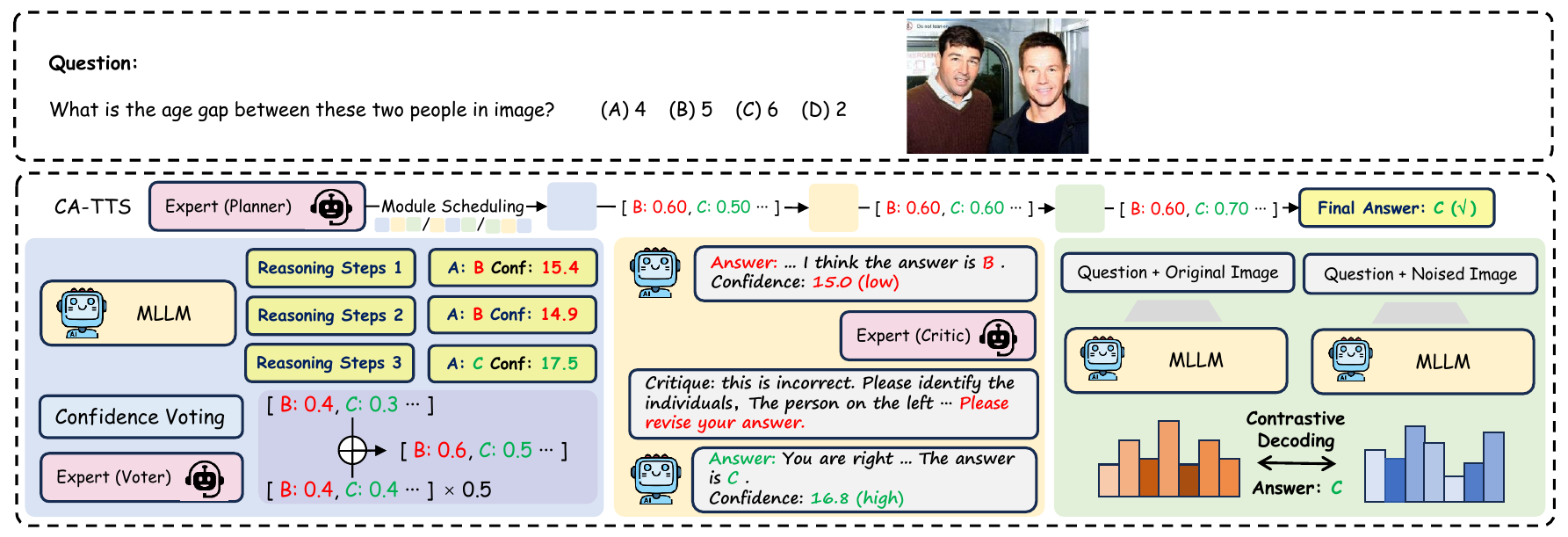}
    \caption{\textbf{A case study of CA-TTS on MMStar.} Our method (bottom) demonstrates a multi-stage, resilient process: an initial error from \textbf{Self-Consistency} (Answer: B) is corrected by \textbf{Self-Reflection} (Answer: C) and confirmed by \textbf{Self-Check}.}
    \label{fig:case_study2}
\end{figure*}

\begin{figure*}[h]
    \centering
    \includegraphics[width=\linewidth]{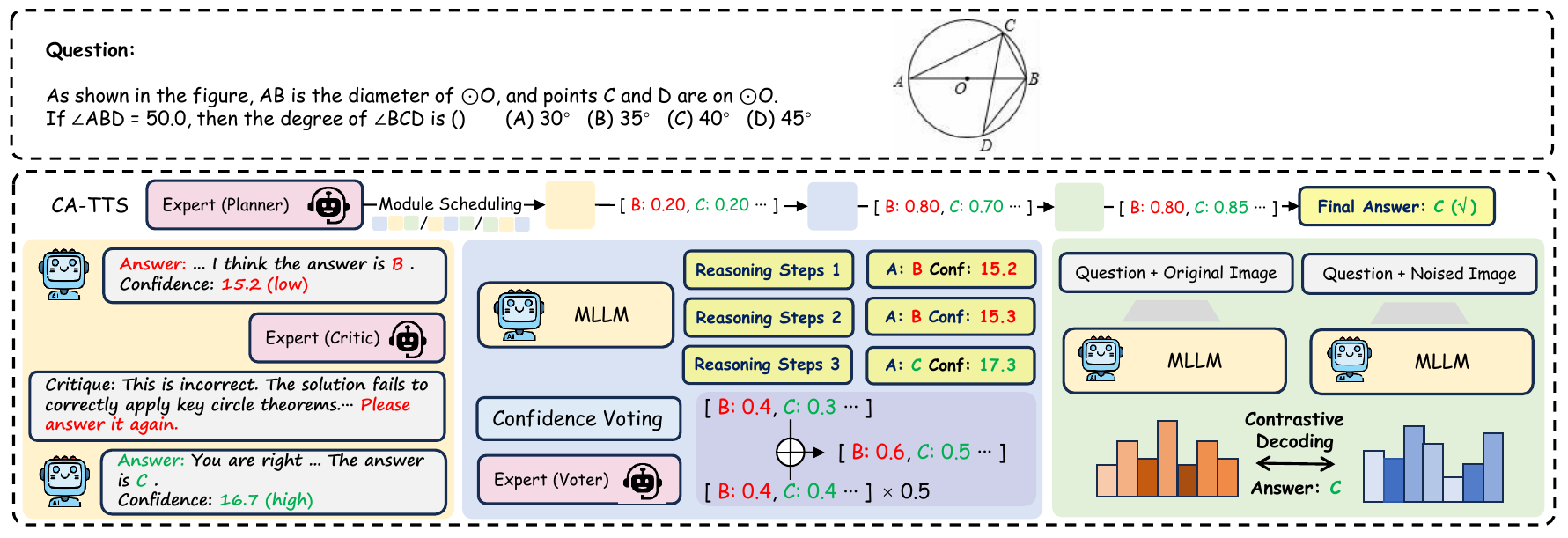}
    \caption{\textbf{A case study of CA-TTS on Math-Vista.} Our method (bottom) demonstrates a multi-stage, resilient process: a error from \textbf{Self-Reflection} and \textbf{Self-Consistency}(Answer: B) is corrected by \textbf{Self-Check} (Answer: C).}
    \label{fig:case_study3}
\end{figure*}

Figures \ref{fig:case_study2} and \ref{fig:case_study3} illustrate additional case studies of our framework. Consistent with the analysis in the main text~\ref{case_study}, our approach exhibits a distinct superiority over tree-based paradigms (e.g., ToT). While ToT is often characterized by a cumbersome and non-decouplable architecture with complex branching logic, it is further limited by single-round verification, leading to an excessive reliance on the inherent capability of the Expert model.

\section{Discussions} 

\noindent\textbf{Justification for Confidence Calculation.} 
In our implementation, we adopt the Normalized Mean Log-Probability (NMLP) as the core metric, consistent with DeepConf~\cite{fu2025deep}. Although we experimented with alternative aggregation strategies, such as \textit{Tail Confidence} (prioritizing final tokens) and \textit{Bottom-Group Confidence} (averaging the lowest probability tokens based on the ``weakest link'' theory), our empirical results consistently favor the global average confidence. We attribute this to the holistic nature of chain-of-thought reasoning, where the validity of the final answer is contingent upon the semantic coherence of the entire reasoning path. Furthermore, local metrics proved overly sensitive to noise, often penalizing high-entropy tokens associated with benign stylistic choices (e.g., synonyms or formatting) rather than factual errors. Consequently, the global average acts as a robust smoothing filter, effectively representing the model's overall certainty regarding the semantic integrity of the output.

\noindent\textbf{Future Prospects.} The vast potential of Test-Time Scaling (TTS) remains largely untapped. In this work, we have empirically demonstrated that confidence awareness acts as a pivotal catalyst in optimizing test-time scaling laws. Looking ahead, we envision extending the principles of our CA-TTS framework to broader paradigms. Specifically, the high-quality reasoning trajectories filtered by our system can serve as gold-standard supervision for \textit{Data Synthesis Frameworks} and \textit{Reinforcement Fine-Tuning}, effectively closing the loop between inference-time scaling and post-training improvements. Furthermore, integrating this dynamic evaluation capability into \textit{Agent-Evolving} systems and \textit{Self-Play} mechanisms holds the promise of enabling models to autonomously refine their strategies. We remain committed to exploring these avenues to further unlock the versatility and impact of TTS in multimodal intelligence.